\documentclass[10pt,twocolumn,letterpaper]{article}

\usepackage{cvpr}              

\usepackage{graphicx}
\usepackage{amsmath}
\usepackage{amssymb}
\usepackage{booktabs}
\usepackage{arydshln}
\usepackage[accsupp]{axessibility}

\usepackage{MnSymbol}
\usepackage{url}
\usepackage{multicol}
\usepackage{multirow}
\usepackage{pifont}
\usepackage{subcaption}
\usepackage[table]{xcolor}
\usepackage{algorithm2e}
\definecolor{mygreen}{rgb}{0,0.5,0}
\definecolor{LightCyan}{rgb}{0.88,1,1}
\usepackage[pagebackref,breaklinks,colorlinks]{hyperref}

\usepackage[capitalize]{cleveref}
\crefname{section}{Sec.}{Secs.}
\Crefname{section}{Section}{Sections}
\Crefname{table}{Table}{Tables}
\crefname{table}{Tab.}{Tabs.}

\def\cvprPaperID{7064} 
\def\confName{CVPR}
\def\confYear{2023}

\newcommand{\cmark}{\ding{51}}%
\newcommand{\xmark}{\ding{55}}%

\begin{document}

\title{Defending Against Patch-based Backdoor Attacks on Self-Supervised Learning}


\author{Ajinkya Tejankar \thanks{Corresponding author {\tt \small <atejankar@ucdavis.edu>}. Work done while interning at Meta AI.} $^{1}$ \quad Maziar Sanjabi $^{2}$ \quad Qifan Wang $^{2}$ \quad Sinong Wang $^{2}$ \quad Hamed Firooz $^{2}$ \\ Hamed Pirsiavash $^{1}$ \quad Liang Tan $^{2}$ \\
$^1$ University of California, Davis \quad $^{2}$ Meta AI\\
}

\maketitle


\begin{abstract}
Recently, self-supervised learning (SSL) was shown to be vulnerable to patch-based data poisoning backdoor attacks. It was shown that an adversary can poison a small part of the unlabeled data so that when a victim trains an SSL model on it, the final model will have a backdoor that the adversary can exploit. This work aims to defend self-supervised learning against such attacks. We use a three-step defense pipeline, where we first train a model on the poisoned data. In the second step, our proposed defense algorithm (PatchSearch) uses the trained model to search the training data for poisoned samples and removes them from the training set. In the third step, a final model is trained on the cleaned-up training set. Our results show that PatchSearch is an effective defense. As an example, it improves a model's accuracy on images containing the trigger from 38.2\% to 63.7\% which is very close to the clean model's accuracy, 64.6\%. Moreover, we show that PatchSearch outperforms baselines and state-of-the-art defense approaches including those using additional clean, trusted data. Our code is available at \textcolor{magenta}{\href{https://github.com/UCDvision/PatchSearch}{https://github.com/UCDvision/PatchSearch}}
\end{abstract}


\section{Introduction}

\begin{figure*}
    \centering
    \includegraphics[width=\textwidth]{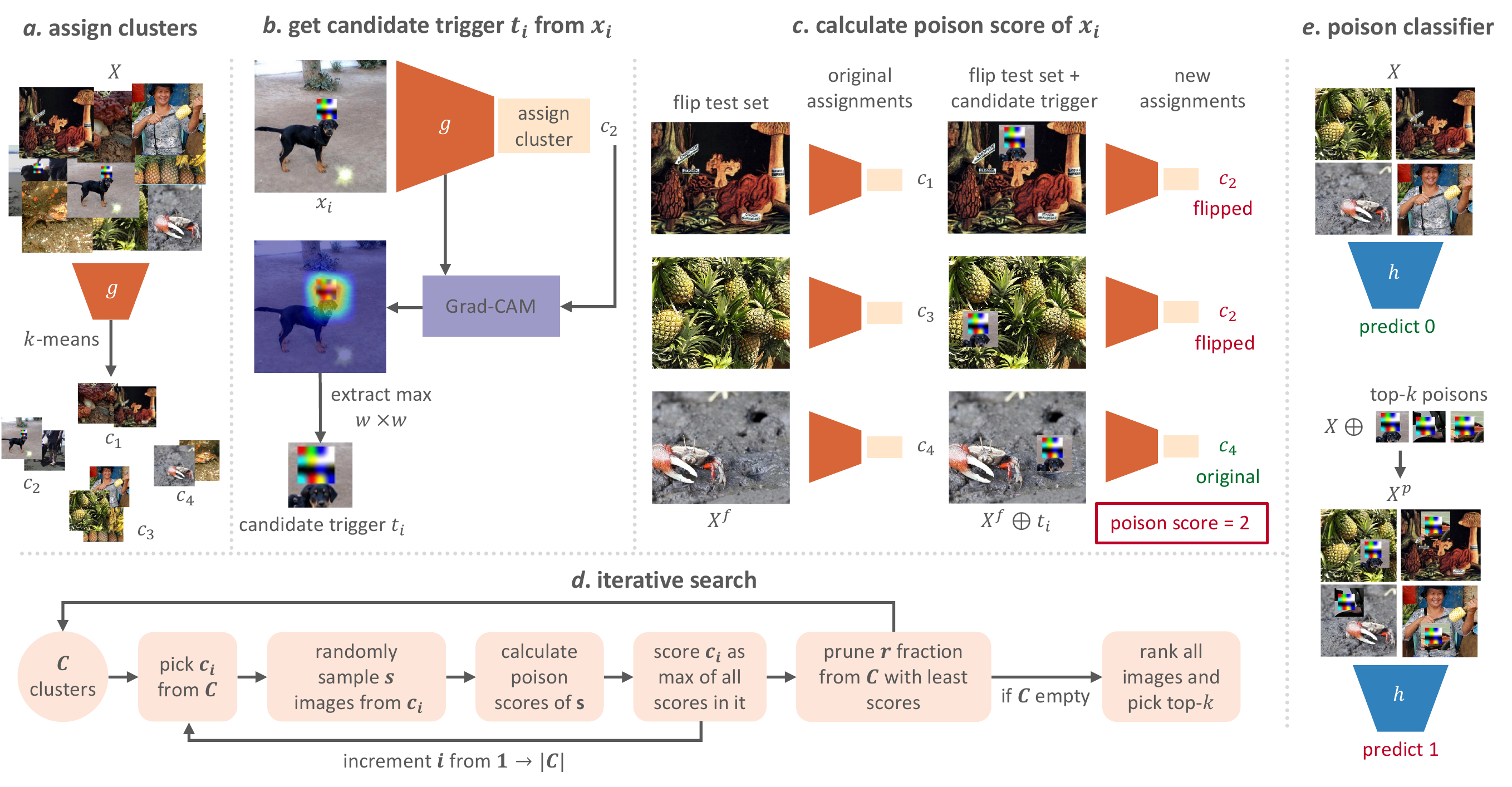}
    \caption{\textbf{Illustration of PatchSearch.} SSL has been shown to group visually similar images in  \cite{abbasi2020compress,soroush2021mean}, and \cite{saha2022backdoor} showed that poisoned images are close to each other in the representation space. Hence, the very first step (a) is to cluster the training dataset. We use clustering to locate the trigger and to efficiently search for poisoned samples. The second step locates the candidate trigger in an image with Grad-CAM (b) and assigns a poison score to it (c). The third step (d) searches for highly poisonous clusters and only scores images in them. This step outputs a few highly poisonous images which are used in the fourth and final step (e) to train a more accurate poison classifier.}
    \label{fig:teaser}
\end{figure*}

Self-supervised learning (SSL) promises to free deep learning models from the constraints of human supervision. It allows models to be trained on cheap and abundantly available unlabeled, uncurated data \cite{goyal2021self}. A model trained this way can then be used as a general feature extractor for various downstream tasks with a small amount of labeled data. However, recent works \cite{saha2022backdoor,carlini2022poisoning} have shown that uncurated data collection pipelines are vulnerable to data poisoning backdoor attacks. 


Data poisoning backdoor attacks on self-supervised learning (SSL) \cite{saha2022backdoor} work as follows. An attacker introduces ``backdoors'' in a model by simply injecting a few carefully crafted samples called ``poisons'' in the unlabeled training dataset. Poisoning is done by pasting an attacker chosen ``trigger'' patch on a few images of a ``target'' category. 
A model pre-trained with such a poisoned data behaves similar to a non-poisoned/clean model in all cases except when it is shown an image with a trigger. Then, the model will cause a downstream classifier trained on top of it to mis-classify the image as the target category. This types of attacks are sneaky and hard to detect since the trigger is like an attacker's secret key that unlocks a failure mode of the model. Our goal in this work is to defend SSL models against such attacks.


While there have been many works for defending supervised learning against backdoor attacks \cite{li2022backdoor}, many of them directly rely upon the availability of labels. Therefore, such methods are hard to adopt in SSL where there are no labels. However, a few supervised defenses \cite{hong2020effectiveness,borgnia2021strong} can be applied to SSL with some modifications. One such defense, proposed in \cite{hong2020effectiveness}, shows the effectiveness of applying a strong augmentation like CutMix \cite{yun2019cutmix}. Hence, we adopt $i$-CutMix \cite{lee2021imix}, CutMix modified for SSL, as a baseline. We show that $i$-CutMix is indeed an effective defense that additionally improves the overall performance of SSL models.

Another defense that does not require labels was proposed in \cite{saha2022backdoor}. While this defense successfully mitigates the attack, its success is dependent on a significant amount of clean, trusted data. However, access to such data may not be practical in many cases. Even if available, the trusted data may have its own set of biases depending on its sources which might bias the defended model. Hence, our goal is to design a defense that does not need any trusted data.

In this paper, we propose PatchSearch, which aims at identifying poisoned samples in the training set without access to trusted data or image labels. One way to identify a poisoned sample is to check if it contains a patch that behaves similar to the trigger. A characteristic of a trigger is that it occupies a relatively small area in an image ($\approx$ 5\%) but heavily influences a model's output. Hence, we can use an input explanation method like Grad-CAM \cite{selvaraju2017grad} to highlight the trigger. Note that this idea has been explored in supervised defenses \cite{doan2020februus,chou2020sentinet}. However, Grad-CAM relies on a supervised classifier which is unavailable in our case. This is a key problem that we solve with $k$-means clustering. An SSL model is first trained on the poisoned data and its representations are used to cluster the training set. SSL has been shown to group visually similar images in  \cite{abbasi2020compress,soroush2021mean}, and \cite{saha2022backdoor} showed that poisoned images are close to each other in the representation space. Hence, we expect poisoned images to be grouped together and the trigger to be the cause of this grouping. Therefore, cluster centers produced by $k$-means can be used as classifier weights in place of a supervised classifier in Grad-CAM. Finally, once we have located a salient patch, we can quantify how much it behaves like a trigger by pasting it on a few random images, and counting how many of them get assigned to the cluster of the patch (Figure \ref{fig:teaser} (c)). A similar idea has been explored in \cite{chou2020sentinet}, but unlike ours, it requires access to trusted data, it is a supervised defense, and it operates during test time.


One can use the above process of assigning poison scores to rank the entire training set and then treat the top ranked images as poisonous. However, our experiments show that in some cases, this poison detection system has very low precision at high recalls. Further, processing all images is redundant since only a few samples are actually poisonous. For instance, there can be as few as 650 poisons in a dataset of 127K samples ($\approx$ 0.5\%) in our experiments. Hence, we design an iterative search process that focuses on finding highly poisonous clusters and only scores images in them (Figure \ref{fig:teaser} (d)). The results of iterative search are a few but highly poisonous samples. These are then used in the next step to build a poison classifier which can identify poisons more accurately. This results in a poison detection system with about $55\%$ precision and about $98\%$ recall in average.

In summary, we propose PatchSearch, a novel defense algorithm that defends self-supervised learning models against patch based data poisoning backdoor attacks by efficiently identifying and filtering out poisoned samples. We also propose to apply $i$-CutMix \cite{lee2021imix} augmentation as a simple but effective defense. Our results indicate that PatchSearch is better than both the trusted data based defense from \cite{saha2022backdoor} and $i$-CutMix. Further, we show that $i$-CutMix and PatchSearch are complementary to each other and combining them results in a model with an improved overall performance while significantly mitigating the attack.

\section{Related Work}
\textbf{Self-supervised learning.} The goal of self-supervised learning (SSL) is to learn representations from uncurated and unlabeled data. It achieves this with a ``pretext task'' that is derived from the data itself without any human annotations \cite{zhang2016colorful,gidaris2018unsupervised,noroozi2016unsupervised,dosovitskiy2014discriminative,wu2018unsupervised}. MoCo \cite{he2020momentum} is a popular SSL method in which we learn by classifying a pair of positives, two augmentations of the same image, against pairs of negatives, augmentations from different images. This is further explored in \cite{chen2020mocov2,chen2020simple,tian2020makes,caron2020unsupervised}. BYOL \cite{grill2020bootstrap} showed that it is possible to train the model without negatives. This was improved in \cite{tejankar2021isd,soroush2021mean}. 

\textbf{Backdoor attacks and defenses for supervised learning.} The goal of these attacks is to cause a model to misclassify images containing an attacker chosen trigger. The attack is conducted by poisoning the training dataset. These attacks were first studied for supervised learning in \cite{gu2017badnets,liu2017trojaning,liu2017neural}. More advanced attacks that are harder to detect have been proposed in \cite{saha2020hidden,shafahi2018poison,turner2018clean}. We specifically study patch based \cite{brown2017adversarial,saha2020role} version of backdoor attacks.

There is a large literature devoted to defending against patch-based attacks on supervised learning \cite{li2022backdoor}. Following are a few important paradigms of defenses. 
(1) \textit{pre-processing the input} \cite{doan2020februus}: remove the trigger from a given input before forwarding it through the model. (2) \textit{trigger synthesis} \cite{wang2019neural}: explicitly identify the trigger and then modify the model to ignore it. (3) \textit{model diagnostics} \cite{kolouri2020universal,huang2019neuroninspect,zheng2021topological}: classify whether a given model is poisoned or not. (4) \textit{model reconstruction} \cite{liu2018fine}: directly modify the model weights such that the trigger cannot have any impact on it. (5) \textit{backdoor suppression} \cite{hong2020effectiveness,borgnia2021strong,li2021anti,huang2022backdoor}: modify the training procedure of the model such that backdoors don't get learned by the model in the first place. We show that a defense from this paradigm in the form of $i$-CutMix augmentation \cite{lee2021imix} is a simple and effective baseline. Note that $i$-CutMix \cite{lee2021imix} is a version of CutMix \cite{yun2019cutmix} that does not need labels. Moreover, strong augmentations like CutMix were shown to be better compared to other backdoor suppression techniques like \cite{hong2020effectiveness} which are based on differential privacy. (6) \textit{training sample filtering} \cite{tran2018spectral,chen2018detecting,zeng2021rethinking,huang2020one}: first identify the poisoned samples, then filter them out of the training dataset, and finally train a new model on the resulting clean dataset. This is also the paradigm of our defense. (7) \textit{testing sample filtering} \cite{chou2020sentinet,gao2019strip}: identify poisoned samples during the inference phase of a model and prevent them from being forwarded through the model. Similar to these works \cite{chou2020sentinet,doan2020februus} we use Grad-CAM \cite{selvaraju2017grad} to locate the trigger, but without access to a supervised classifier we calculate Grad-CAM with a $k$-means based classifier. Further, similar to ours \cite{chou2020sentinet} pastes the located trigger on a few images to calculate its poisonousness, but they assume that these images are from a trusted source. Moreover, unlike \cite{doan2020februus,chou2020sentinet}, ours is a train time filtering defense. 

Finally, above supervised defenses cannot be directly adopted to self-supervised learning. For instance, consider Activation Clustering \cite{chen2018detecting} which uses the idea that poisoned images cluster together. However, their full idea is to perform 2-way clustering within each class, and consider a class as poisoned if the 2-way clustering is imbalanced. In the absence of labels, this defense is not applicable in SSL.

\textbf{Data poisoning backdoor attacks and their defenses for self/weakly supervised learning.} These attacks were first studied for self-supervised learning recently in \cite{saha2022backdoor}. The work also proposed a defense but it requires access to a non-trivial amount of trusted data. Unlike this defense, PatchSearch does not require any trusted data. Further, \cite{carlini2022poisoning} showed that weakly-supervised models like CLIP \cite{radford2021learning} that use uncurated data can be attacked.

\section{Threat Model}

We adopt the backdoor attack threat model proposed in \cite{saha2022backdoor}. The attacker is interested in altering the output of a classifier trained on top of a self-supervised model for a downstream task. The attacker's objective is twofold. First, insert a backdoor into the self-supervised model such that the downstream classifier mis-classifies an incoming image as the target category if it contains the attacker chosen trigger. Second, hide the attack by making sure that the accuracy of the downstream classifier is similar to a classifier trained on top of a clean model if the image does not contain the trigger. The attacker achieves these objectives by ``data poisoning'' where a small trigger patch, fixed beforehand, is pasted on a few images of the target category which is also chosen by the attacker. The hypothesis is that the SSL method associates the trigger with the target category resulting in a successful attack. Given this attack, the defender's goal is to detect and neutralize it without affecting the overall performance of the model. The defender must do this without knowing the trigger or the target category and without any access to trusted data or image labels.

\section{Defending with PatchSearch}

The goal of PatchSearch is to identify the poisonous samples in the training set and filter them out. We do this by first training an SSL model on the poisonous data and using it to identify the poisonous samples. The cleaned up training data is then used to train a final, backdoor-free model. Note that the architecture and the SSL method used for the final model need not be the same as the model used during defense. In fact, we show that using an easy to attack model during defense allows PatchSearch to find the poisons more easily. There are four components of PatchSearch as illustrated in Figure \ref{fig:teaser}. (1) Cluster the dataset. (2) Assign a poison score to a given image. (3) Efficiently search for a few but highly poisonous samples. (4) Train a poison classifier to find poisons with high recall. 

We are given an unlabeled dataset $X$ with images $\{x_i\}_{i=1}^{N}$. First, we cluster the representations of $X$ into $l$ clusters with $k$-means to get cluster centers and cluster assignments $(x_i, y_i)$ where $y_i \in \{c_1 ... c_l\}$ is the cluster label (Figure \ref{fig:teaser} (a)). The resulting clusters are used for locating candidate triggers and to efficiently search for poisons. 


\textbf{Scoring an image.} The goal is to quantify the poisonousness of a given image by assigning a poison score to it. The steps for this are described next and illustrated in Figures \ref{fig:teaser} (b) and \ref{fig:teaser} (c). We use the cluster centers as classification weights by calculating cosine similarity between $x_i$ and cluster centers, and use $y_i$ as the label to calculate Grad-CAM for $x_i$. The resulting heatmap is used to extract a candidate trigger $t_i$ which is a $w \times w$ patch that contains the highest heatmap values. We then choose a set of images called flip test set $X^f$, paste $t_i$ on those, forward through the model and get their cluster assignments. Finally, the poison score of $t_i$ or $x_i$ is the number of images in $X^f$ whose cluster assignments flipped to that of $x_i$. We ensure that the flip test set is difficult to flip by sampling a few images per cluster that are closest to their respective cluster centers. Note that all flip set images do not need to be clean.

\textbf{Iterative search.} In this phase, the goal is to efficiently find a few but highly poisonous images. Hence, instead of assigning poison scores to the entire set $X$, we propose a greedy search strategy that focuses on scoring images from poisonous clusters. Concretely, we process $s$ random samples per cluster in an iteration (Figure \ref{fig:teaser} (d)). At the end of an iteration, each cluster is assigned a poison score as the highest poison score of the images in it. Since poisoned images cluster together, just a single iteration should give us a good estimate of clusters that are likely to contain poisons. We use this insight to focus on highly poisonous cluster by removing (pruning) a fraction $r$ of the clusters with least poison scores in each iteration. The algorithm stops when there are no more clusters to be processed. Finally, all scored images are ranked in the decreasing order of their scores and the top-$k$ ranked images are used in the next step. 


\textbf{Poison classification.} Here, the goal is to find as many poisons as possible. We do this by building a classifier $h$ to identify the top-$k$ ranked patches found in the previous step (Figure \ref{fig:teaser} (e)). Specifically, given the training dataset $X$ and a set $P$ of $k$ patches, we construct poisoned training set $X^p$ by randomly sampling a patch $p_j$ from $P$ and pasting it at a random location on $x_i$. Samples from $X$ are assigned label ``$0$'' (non-poisonous) while samples from $X^p$ are assigned the label ``$1$'' (poisonous). However, this labeling scheme is noisy since there are poisoned images in $X$ which will be wrongly assigned the label $0$. One way to reduce noise is to not include images in $X$ with high poison scores. Another way is to prevent the model from overfitting to the noise by using strong augmentations and early stopping. However, early stopping requires access to validation data which is unavailable. Hence, we construct a proxy validation set by using the top-$k$ samples with label $1$ (poisonous) and bottom-$k$ samples with label $0$ (non-poisonous). Note that these samples are excluded from the training set. We stop training the classifier if F1 score on the proxy validation set does not change for 10 evaluation intervals. Finally, the resulting binary classifier is applied to $X$ and any images with ``$1$'' (poisonous) label are removed from the dataset.


\section{Experiments}

We follow the experimentation setup proposed in \cite{saha2022backdoor}.

\textbf{Datasets.} We use ImageNet-100 dataset \cite{tian2020contrastive}, which contains images of 100 randomly sampled classes from the 1000 classes of ImageNet \cite{imagenet}. The train set has about 127K samples while the validation set has 5K samples.

\textbf{Architectures and self-supervised training.} In addition to ResNet-18 \cite{he2016deep}, we also conduct experiments on ViT-B \cite{dosovitskiy2021an} with MoCo-v3 \cite{chen2021empirical} and MAE \cite{he2022masked}. Unlike \cite{saha2022backdoor}, we only consider ResNet-18 with MoCo-v2 \cite{chen2020mocov2} and BYOL \cite{grill2020bootstrap} since we find that older methods like Jigsaw \cite{noroozi2016unsupervised} and RotNet \cite{gidaris2018unsupervised} have significantly worse clean accuracies compared to MoCo-v2 and BYOL. For the code and parameters, we follow MoCo-v3 \cite{chen2021empirical} for ViT-B, and \cite{saha2022backdoor} for ResNet-18. Finally, implementation of $i$-CutMix and its hyperparameter values follow the official implementation from \cite{lee2021imix}.

\textbf{Evaluation.} We train a linear layer on top of a pre-trained models with 1\% of randomly sampled, clean and labeled training set. To account for randomness, we report results averaged over 5 runs with different seeds. We use the following metrics over the validation set to report model performance: accuracy (Acc), count of False Positives (FP) for the target category, and attack success rate (ASR) as the percentage of FP ($FP*100/4950$). Note that maximum FP is 4950 (50 samples/category $\times$ 99 incorrect categories). The metrics are reported for the validation set with and without trigger pasted on it, referred to as Patched Data and Clean Data. See Section \ref{sec:appendix_implementation_details} of the Appendix for more details. 

\begin{figure}
    \centering
    \includegraphics[width=\columnwidth]{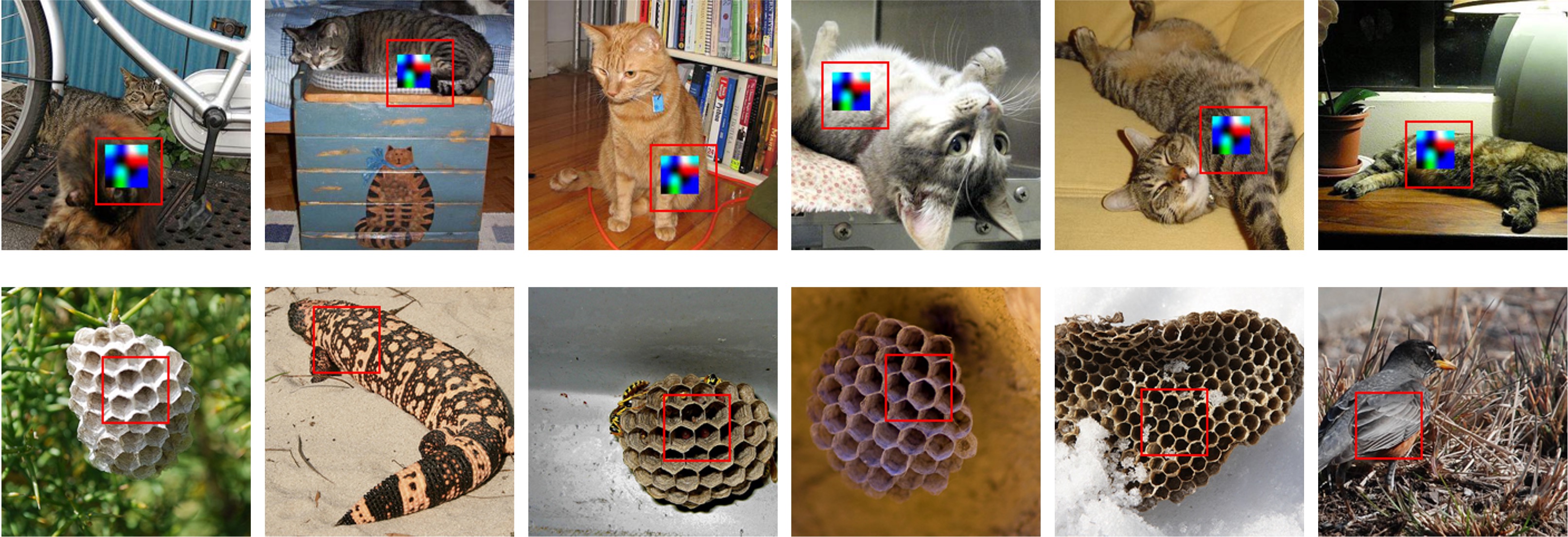}
    \caption{\textbf{Top ranked poisonous and non-poisonous images.} We visualize a few top ranked poisonous (1st row) and non-poisonous (2nd row) images found by PatchSearch for BYOL, ResNet-18, poison rate 0.5\% and target category Tabby Cat.} 
    \label{fig:top_tp_fp_poisonous_images}
\end{figure}

\textbf{Attack.} We consider 10 different target categories and trigger pairs used in \cite{saha2022backdoor}. Unless mentioned otherwise, all results are averaged over the 10 target categories. Further, following \cite{saha2022backdoor}, we consider 0.5\% and 1.0\% poison injection rates which translate to 50\% and 100\% images being poisoned for a target category. Poisons are generated according to \cite{saha2022backdoor,backdoor-code} (see Figure \ref{fig:top_tp_fp_poisonous_images} for an example). The pasted trigger size is 50x50 ($\approx$ 5\% of the image area) and it is pasted at a random location in the image while leaving a margin of 25\% on all sides of the image.

\textbf{PatchSearch.} Grad-CAM implementation is used from \cite{jacobgilpytorchcam}. Unless mentioned, we run the PatchSearch with following parameters: number of clusters $l$ = 1000, patch size $w$ = 60, flip test set size = 1000, samples per cluster $s$ = 2, and clusters to prune per iteration $r$ = 25\%. This results in about 8K (6\%) training set images being scored. The architecture of the poison classifier is based on ResNet-18 \cite{he2016deep} but uses fewer parameters per layer. We use top-$20$ poisons to train the classifier, and remove top 10\% of poisonous samples to reduce noise in the dataset. We use strong augmentations proposed in MoCo-v2 \cite{chen2020mocov2} and randomly resize the patch between 20x20 and 80x80 (where input is $w \times w$ and default $w = 60$). The classifier is evaluated every 20 iterations with max 2K iterations, and training is stopped if F1 score on the proxy val set does not change for 10 evaluations. Finally, in order to improve the stability of results, we train 5 different classifiers and average their predictions. See Section \ref{sec:appendix_implementation_details} in Appendix for more details. 


\begin{table}
    \centering
    \scalebox{0.9}{
    \begin{tabular}{@{} lcccc @{}}
    \toprule
        & top-$20$ & Rec. & Prec. & Total \\
        Model & Acc (\%) & (\%) & (\%) & Rem. \\
        \midrule
        BYOL, ResNet-18, 0.5\% & 99.5 & 98.0 & 58.8 & 1114.6 \\
        MoCo-v3, ViT-B, 0.5\%  & 96.5 & 98.8 & 48.3 & 1567.0 \\
        MoCo-v3, ViT-B, 1.0\%  & 97.5 & 99.0 & 59.5 & 2368.3 \\
    \bottomrule
    \end{tabular}
    }
    \caption{\textbf{Poison detection results.} We find that iterative search has high \textit{top-$20$ Acc} which is the accuracy of finding poisons in top-$20$ ranked images. In the right 3 columns, we show the effectiveness of the poison classification model at filtering out the poisons. \textit{Rec.} is recall, \textit{Prec.} is precision, while \textit{Total Rem.} is the total number of samples removed by the classifier. We can see that the classifier can identify most of the poisons since recall is $\approx$ 98\%. Note that removing a small number of clean images does not significantly damage the final SSL model as shown in Table \ref{tab:main_defense_results}.}
    
    \label{tab:defense_ranking_results}
\end{table}

\begin{table*}[h]
    \centering
    \renewcommand{\arraystretch}{1.2}
    \scalebox{0.90}{
    \begin{tabular}{ l c ccc|crr c ccc|crr }
    \toprule
    \multirow{2}{*}{Model Type} & & \multicolumn{3}{c}{Clean Data} & \multicolumn{3}{c}{Patched Data} & & \multicolumn{3}{c}{Clean Data} & \multicolumn{3}{c@{}}{Patched Data} \\
    \cmidrule(lr){3-5} \cmidrule(lr){6-8} \cmidrule(lr){10-12} \cmidrule(l){13-15}
    & & Acc & FP & ASR & Acc & FP & ASR & & Acc & FP & ASR & Acc & FP & ASR \\
    \midrule
    \textit{ViT-B} & & \multicolumn{6}{c}{\textit{MoCo-v3, poison rate 0.5\%}} & & \multicolumn{6}{c}{\textit{MoCo-v3, poison rate 1.0\%}} \\
    \cmidrule(lr){1-1} \cmidrule(lr){3-8} \cmidrule(l){10-15}
    Clean      & \cellcolor{white} & 70.5 & 18.5 & 0.4 & 64.6 & 27.2 & 0.5 & \cellcolor{white} & 70.5 & 18.5 & 0.4 & 64.7 & 27.2 & 0.5 \\
    \rowcolor{gray!15}
    Clean + $i$-CutMix & \cellcolor{white} & 75.6 & 15.6 & 0.3 & 74.4 & 14.6 & 0.3 & \cellcolor{white} & 75.6 & 15.6 & 0.3 & 74.4 & 14.6 & 0.3 \\
    Backdoored & \cellcolor{white} & 70.6 & 17.4 & 0.4 & 46.9 & {\bf 1708.9} & 34.5 & \cellcolor{white} & 70.7 & 14.8 & 0.3 & 37.7 & {\bf 2535.9} & 51.2 \\
    \rowcolor{gray!15}
    Backdoored + $i$-CutMix & \cellcolor{white} & 75.6 & 14.9 & 0.3 & 72.2 & 242.2 & 4.9 & \cellcolor{white} & 75.5 & 12.5 & 0.3 & 70.2 & 434.2 & 8.8 \\
    PatchSearch & \cellcolor{white} & 70.2 & 23.1 & 0.5 & 64.5 & 39.8 & 0.8 & \cellcolor{white} & 70.1 & 43.4 & 0.9 & 63.7 & 76.0 & 1.5 \\
    \rowcolor{gray!15}
    PatchSearch + $i$-CutMix & \cellcolor{white} & 75.2 & 19.7 & 0.4 & 74.2 & {\bf 19.0} & 0.4 & \cellcolor{white} & 75.0 & 39.0 & 0.8 & 74.0 & {\bf 41.4} & 0.8 \\
    
    \midrule
    
    \textit{ResNet-18} & & \multicolumn{6}{c}{\textit{MoCo-v2, poison rate 0.5\%}} & & \multicolumn{6}{c}{\textit{BYOL, poison rate 0.5\%}} \\
    \cmidrule(lr){1-1} \cmidrule(lr){3-8} \cmidrule(l){10-15}
    Clean      & \cellcolor{white} & 49.7 & 34.0 & 0.7 & 46.5 & 35.4 & 0.7 & \cellcolor{white} & 65.7 & 20.6 & 0.4 & 60.6 & 24.8 & 0.5 \\
    \rowcolor{gray!15}
    Clean + $i$-CutMix & \cellcolor{white} & 55.9 & 26.3 & 0.5 & 54.3 & 27.1 & 0.5 & \cellcolor{white} & 65.8 & 20.5 & 0.4 & 63.7 & 19.6 & 0.4 \\
    Backdoored & \cellcolor{white} & 50.0 & 27.2 & 0.5 & 40.9 & {\bf 703.3} & 14.2 & \cellcolor{white} & 66.5 & 18.6 & 0.4 & 35.3 & {\bf 2079.6} & 42.0 \\
    \rowcolor{gray!15}
    Backdoored + $i$-CutMix & \cellcolor{white} & 55.4 & 24.4 & 0.5 & 53.0 & 124.0 & 2.5 & \cellcolor{white} & 67.0 & 18.5 & 0.4 & 61.0 & 365.3 & 7.4 \\
    PatchSearch   & \cellcolor{white} & 49.7 & 33.1 & 0.7 & 46.3 & 37.2 & 0.8 & \cellcolor{white} & 66.4 & 22.9 & 0.5 & 61.5 & 33.4 & 0.7 \\
    \rowcolor{gray!15}
    PatchSearch + $i$-CutMix & \cellcolor{white} & 55.3 & 30.6 & 0.6 & 53.8 & {\bf 29.2} & 0.6 & \cellcolor{white} & 66.8 & 22.4 & 0.5 & 64.5 & {\bf 22.5} & 0.5 \\
    \bottomrule
    \end{tabular}
    }
    \caption{\textbf{Defense results.} We compare clean, backdoored, and defended models in different configurations. \textit{Acc} is the accuracy on validation set, \textit{FP} is the number of false positives for the target category, and \textit{ASR} is FP as percentage. All results are averaged over 5 evaluation runs and 10 categories. We observe the following: (1) Presence of $i$-CutMix significantly boosts clean accuracy for models trained with MoCo, but this boost is small for BYOL models. (2) While $i$-CutMix reduces the effectiveness of attacks in all cases, there is still room for improvement. (3) PatchSearch alone can significantly mitigate backdoor attacks. (4) \textit{patched accuracies} of models defended with PatchSearch + $i$-CutMix are very close to clean models in all cases. See Section \ref{sec:appendix_per_category_results} for detailed, category-wise results.}
    \label{tab:main_defense_results}
\end{table*}



\begin{table}
    \centering
    \scalebox{0.9}{
    \begin{tabular}{@{}lr cccr@{}}
    \toprule
        \multirow{2}{*}{Model} &  \multirow{2}{*}{\shortstack{Trusted\\Data}} & \multicolumn{2}{c}{Clean Data} & \multicolumn{2}{c@{}}{Patched Data} \\
        \cmidrule(lr){3-4} \cmidrule(l){5-6}
        & & Acc & FP & Acc & FP \\
        \midrule
        Clean & 100\% & 49.9 & 23.0 & 47.0 & 22.8 \\
        Backdoored & 0\% & 50.1 & 26.2 & 31.8 & 1683.2 \\
        KD Defense & 25\% & 44.6 & 34.5 & 42.0 & 37.9 \\
        KD Defense & 10\% & 38.3 & 40.5 & 35.7 & 44.8 \\
        KD Defense & 5\% & 32.1 & 41.0 & 29.4 & 53.7 \\
        PatchSearch & 0\% & 49.4 & 40.1 & 45.9 & 50.3 \\
    \bottomrule
    \end{tabular}
    }
    \caption{\textbf{Comparison with trusted-data defense}. We compare PatchSearch with the trusted data and knowledge distillation (KD) defense \cite{saha2022backdoor}. For a fair comparison, we use the setting from \cite{saha2022backdoor} without $i$-CutMix: ResNet-18, MoCo-v2, and poison rate 1.0\%. All results except the last row are from \cite{saha2022backdoor}. We find that despite not relying on any trusted data, our defense can suppress the attack to the level of KD + 5\% trusted data. Moreover, our model has significantly higher accuracies: 49.4\% vs 32.1\% on clean data and 45.9\% vs 29.4\% on patched data. Compared to KD + 25\% trusted data, our model has higher accuracies with only slightly higher FP.}
    \label{tab:defense_comparison}
\end{table}


\subsection{Main Results}

\textbf{Poison detection results of PatchSearch.} We report the results of poison detection for PatchSearch at different stages in Table \ref{tab:defense_ranking_results}. We can see that at the end of iterative search, most of the top-$20$ images are indeed poisonous. We can also see that poison classification leads to very high recall with acceptable precision. Finally, we visualize a few top ranked poisonous and non-poisonous images found with PatchSearch in Figure \ref{fig:top_tp_fp_poisonous_images}. 

\textbf{Effectiveness of PatchSearch and $i$-CutMix.} Table \ref{tab:main_defense_results} lists the results of using $i$-CutMix and PatchSearch for defending against backdoor attacks on 4 different settings. Note that we make it easier for PatchSearch to find poisons by training an easy to attack model during defense, and since $i$-CutMix can suppress the attack, we use models trained without it during PatchSearch.
Similarly, we find that the attack is not strong enough to be detectable by PatchSearch for ResNet-18 and MoCo-v2 models (Table \ref{tab:effect_of_k}).
Therefore, we use the dataset filtered by PatchSearch with ViT-B, MoCo-v3 models. Our results indicate that $i$-CutMix not only improves the clean accuracy of the models but also reduces the attack effectiveness. However, $i$-CutMix cannot fully remove the effect of poisoning and it suppresses the learning of backdoor without explicitly identifying poisons. Here, PatchSearch is a good alternative, which not only significantly suppresses the attack, but also identifies poisoned samples that can be verified by the defender to detect an attack. This may have value in some applications where tracking the source of poisoned samples is important. Finally, the two defenses are complementary to each other, and their combination results in a model that behaves very close to a clean model. PatchSearch removes most poisoned samples while $i$-CutMix suppresses the effect of some remaining ones. Detailed results including per category results, and mean and variance information can be found in Section \ref{sec:appendix_per_category_results} in Appendix. 

\textbf{Comparison with trusted data defense.} Table \ref{tab:defense_comparison} compares PatchSearch to the trusted data defense from \cite{saha2022backdoor}. Their defense trains a student model from a backdoored teacher with knowledge distillation (KD) \cite{abbasi2020compress} on poison-free, trusted data. However, trusted data can be difficult to acquire in SSL and may have its own set of biases. The results in Table \ref{tab:defense_comparison} show that PatchSearch offers a good trade-off between accuracy and FP despite not using any trusted data. Compared to KD with 25\% trusted data, which may not be practical in many cases, PatchSearch is better than KD by $\approx 4$ points on clean and patched accuracies at the cost of a slightly higher FP.



\textbf{Finetuning as a defense.} Since the downstream task data is clean and labeled, a very simple defense strategy would be to finetune SSL models on it instead of only training a linear layer. Hence, we compare finetuning with our defense in Table \ref{tab:effect_of_finetuning}. We find that the attack with linear probing increases the FP from 17 to 1708 while the attack with finetuning the whole model increases the FP from 18 to 685. We see that 685 is still much higher than 18, so the attack is still strong with finetuning the whole model. Obviously, one can do finetuning with a very large learning rate to forget the effect of poisoning, but that reduces the accuracy of the model due to overfitting to the downstream task and forgetting the pretraining. Further, with finetuning, different model weights need to be stored for different downstream tasks instead of a single general model, which may not be desirable. Moreover, in settings like few shot learning, the available data can be too little for effective finetuning. Finally, the results also show that all three defenses, PatchSearch, $i$-CutMix, and finetuning, are complementary to each other and can be combined to get a favorable trade-off between accuracy (Acc) and FP (last row of Table \ref{tab:effect_of_finetuning}).


\begin{table}
    \centering
    \scalebox{0.90}{
    \begin{tabular}{@{}l cccr@{}}
    \toprule
         & \multicolumn{2}{c}{Clean Data} & \multicolumn{2}{c}{Patched Data} \\
        \cmidrule(lr){2-3} \cmidrule(l){4-5}
        Evaluation & Acc & FP & Acc & FP \\
        \midrule
        \multicolumn{5}{l}{\textit{Backdoored}} \\
        Linear & 70.6 & 17.4 & 46.9 & 1708.9 \\
        Finetuned & 72.2 & 18.2 & 63.3 & 685.8 \\
        \cmidrule{1-5}
        \multicolumn{5}{l}{\textit{PatchSearch + $i$-CutMix}} \\
        Linear & 75.2 & 19.7 & 74.2 & 19.0 \\
        Finetuned & 78.1 & 16.9 & 76.6 & 16.8 \\
    \bottomrule
    \end{tabular}
    }
    \caption{\textbf{Finetuning as defense}. We explore finetuning an entire model on the downstream dataset as a defense. While not as good as PatchSearch, we find that finetuning can mitigate the attack. However, the model loses its appeal as a general-purpose backbone and becomes specific to a single downstream task. Finally, note that finetuning is complementary to PatchSearch and $i$-CutMix. Setting: ViT-B, MoCo-v3, and poison rate 0.5\%.}
    \vspace{-.1in}
    \label{tab:effect_of_finetuning}
\end{table}

\textbf{Backdoored MAE vs. defended MoCo-v3.} The results from \cite{saha2022backdoor} indicate that MAE is robust against backdoor attacks. However, there are a few confounding factors such as finetuning and differences in model architecture which need more investigation. We compare finetuned ViT-B models trained with MoCo-v3 and MAE in Table \ref{tab:moco_v3_vs_mae}. We find that MoCo-v3 defended with PatchSearch and $i$-CutMix is better than MAE both in terms of Acc and FP for 1\% labeled finetuning data, but MAE quickly catches up in the 10\% regime. It is possible that these observations are specific to our dataset or model training setup. Hence, we show a similar pattern even for officially pre-trained \cite{mocov3modelsandcode,maemodelsandcode} (on ImageNet-1000) models in Table \ref{tab:moco_v3_vs_mae_official_finetune}.

\begin{table}
    \centering
    \scalebox{0.9}{
    \begin{tabular}{@{} l cccr @{}}
    \toprule
        &  \multicolumn{2}{c}{Clean Data} & \multicolumn{2}{c}{Patched Data} \\
        \cmidrule(lr){2-3} \cmidrule(l){4-5}
        Method & Acc & FP & Acc & FP \\
        \midrule
        \multicolumn{5}{l}{\textit{Finetuned with 1\% labeled data}} \\
        MAE & 65.7 & 18.7 & 53.8 & 97.6 \\
        MoCo-v3 & 78.2 & 20.2 & 76.8 & 17.1 \\
        \midrule
        \multicolumn{5}{l}{\textit{Finetuned with 10\% labeled data}} \\
        MAE & 84.0 & 9.5 & 75.6 & 40.1 \\
        MoCo-v3 & 84.9 & 11.5. & 83.0 & 10.8 \\
    \bottomrule
    \end{tabular}
    }
    \caption{\textbf{Backdoored MAE vs. defended MoCo-v3}. While the inherent robustness of MAE to backdoor attacks \cite{saha2022backdoor} is appealing, we show that a properly defended MoCo-v3 model has better clean accuracy and lower FP compared with a backdoored MAE. Setting: ViT-B, poison rate 0.5\%, and average of 4 target categories (Rottweiler, Tabby Cat, Ambulance, and Laptop).}
    \label{tab:moco_v3_vs_mae}
\end{table}


\begin{table}
    \centering
    \scalebox{0.9}{
    \begin{tabular}{@{} llll c @{}}
    \toprule
        & \multicolumn{4}{c}{Finetuning Data} \\
        Method & 1\% & 10\% & 100\% & 100\%\\
        \midrule
        MAE & 51.5 & 73.0 & \textbf{83.5} & \textbf{83.6} \cite{he2022masked} \\
        MoCo-v3 & \textbf{64.5} & \textbf{74.2} & 83.0 & 83.2 \cite{he2022masked} \\
    \bottomrule
    \end{tabular}
    }
    \caption{\textbf{MoCo-v3 is better than MAE in finetuning on less data.} It is possible that observations in Table \ref{tab:moco_v3_vs_mae} are specific to our dataset or model training setup. Hence, we show that finetuning official, unlabeled ImageNet-1K pre-trained, ViT-B models for MAE and MoCo-v3 still results in MoCo-v3 outperforming MAE in the 1\% regime. However, MAE catches up in 10\% and 100\% regimes. Note that only this table uses the ImageNet-1K dataset.}
    \label{tab:moco_v3_vs_mae_official_finetune}
\end{table}

\textbf{Attack on a downstream task.} Following \cite{saha2022backdoor}, so far we have only considered the case where evaluation dataset is a subset of the training dataset. However, SSL models are intended to be used for downstream datasets that are different from the pre-training dataset. Hence, we show in Table \ref{tab:proper_downstream} that it is possible to attack one of the categories used in the downstream task of Food101 \cite{food101} classification. See Section \ref{sec:appendix_additional_ablations} and Table \ref{tab:proper_downstream_appendix} of Appendix for more details. 

\begin{table}
    \centering
    \scalebox{0.9}{
    \begin{tabular}{@{}l rrrr@{}}
    \toprule
        \multirow{2}{*}{\shortstack{Model}} & \multicolumn{2}{c}{Clean Data} & \multicolumn{2}{c@{}}{Patched Data} \\
        \cmidrule(lr){2-3} \cmidrule(l){4-5}
        & Acc & FP & Acc & FP \\
        \midrule
        Clean & 47.6 & 33.8 & 39.4 & 28.0 \\
        Backdoored & 47.9 & 27.5 & 12.4 & 3127.3 \\
        PatchSearch & 47.4 & 35.0 & 40.7 & 32.3 \\
    \bottomrule
    \end{tabular}
    }
    \caption{\textbf{Attack on downstream Food101 classification}. We choose a target category from Food101 \cite{food101} dataset, poison 700 of its images, and add the poisoned images to the pre-training dataset (ImageNet-100). The models are evaluated on Food101 that has 5050 val images in total (50 val images per class $\times$ 101 classes). We find that the attack is successful and also that PatchSearch can defend against the attack. Setting: BYOL \& ResNet-18. See Table \ref{tab:proper_downstream_appendix} of Appendix for detailed results. 
    }
    \label{tab:proper_downstream}
\end{table}

\subsection{Ablations}

\begin{figure}
    \centering
    \includegraphics[width=\columnwidth]{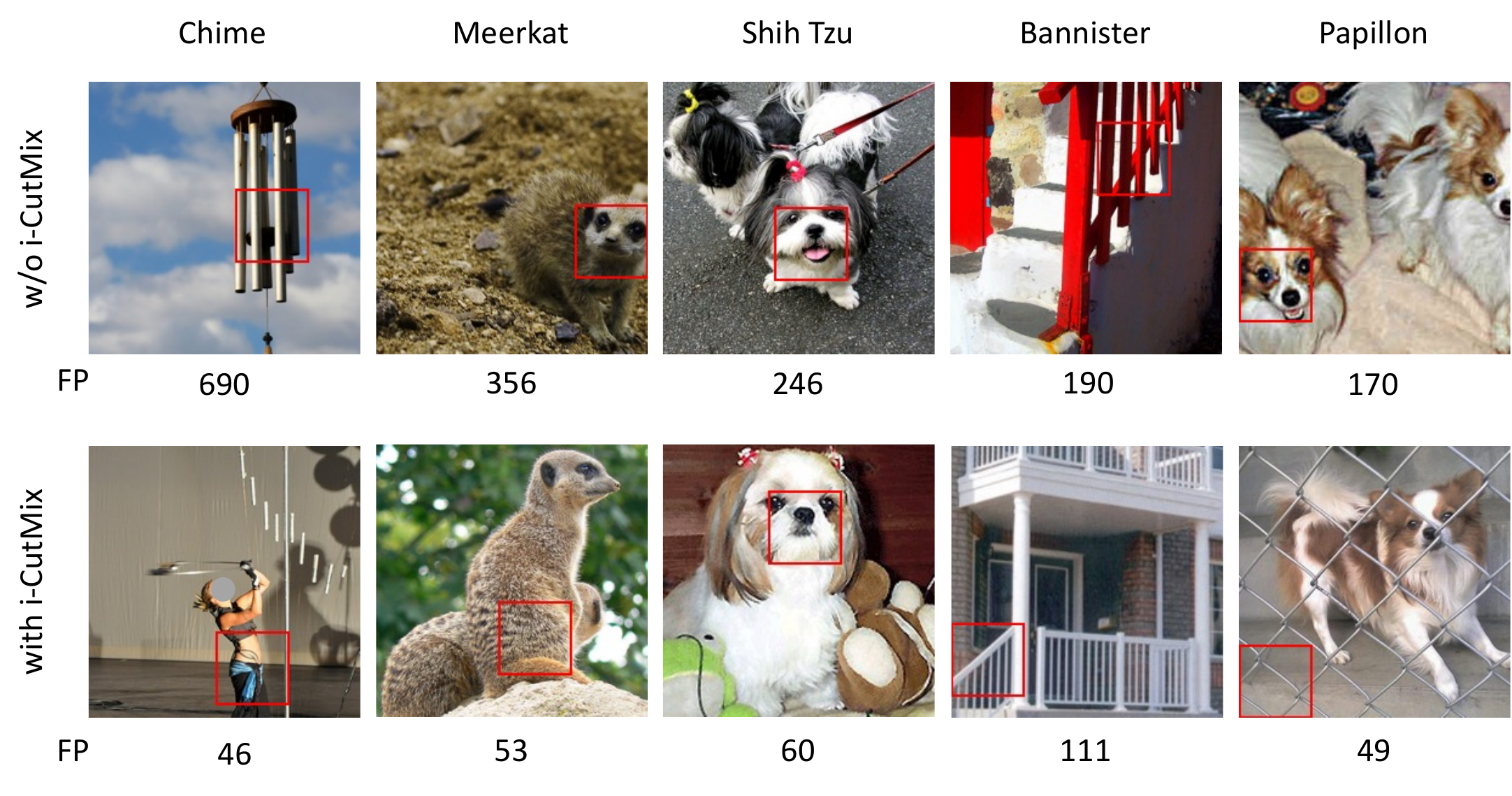}
    \caption{\textbf{Effect of $i$-CutMix on potency of patches.} We hypothesize that $i$-CutMix encourages the model to use global features rather than focusing on small regions. To explore this, we use PatchSearch as an interpretation algorithm and analyze the impact of $i$-CutMix on clean patches. We use a \textit{clean} model to calculate FP of all clean training images with PatchSearch (no iterative search) and use the resulting FP as potency score. FP for each category is calculated as the max FP of all patches in it (shown below images). We find that $i$-CutMix suppresses the impact of individual patches (resulting in smaller potency score), which is aligned with our hypothesis. The total potency (sum of FP) of patches from all categories is reduced by 71.3\% for BYOL, ResNet-18 and 45.4\% for MoCo-v3, ViT-B. We show images for five random categories with and without $i$-CutMix (BYOL, ResNet-18).}
    \label{fig:icutmix_effect}
\end{figure}

\textbf{$i$-CutMix.} It is interesting that a method as simple as $i$-CutMix can be a good defense. Therefore, we attempt to understand which components of $i$-CutMix make it a good defense. $i$-CutMix has two components: creating a composite image by pasting a random crop from a random image, and replacing the original one-hot loss with a weighted one. The weights are in proportion to the percentages of the mixed images. We observe the effect of these components in Table \ref{tab:icutmix_analysis}. We can also get an insight into the working of $i$-CutMix by using PatchSearch as an interpretation algorithm. See Figure \ref{fig:icutmix_effect}. 



\begin{table}
    \centering
    \scalebox{0.90}{
    \begin{tabular}{@{}l cccr@{}}
    \toprule
        & \multicolumn{2}{c}{Clean Data} & \multicolumn{2}{c}{Patched Data} \\
        \cmidrule(lr){2-3} \cmidrule(l){4-5}
        Experiment & Acc & FP & Acc & FP \\
        \midrule
        No $i$-CutMix & 70.5 & 24.8 & 42.9 & 1412.4 \\
        $i$-CutMix & 75.6 & 28.0 & 70.5 & 268.4 \\
        \quad + remove weighted loss & 71.4 & 26.4 & 68.7 & 104.2 \\
        \quad \quad + paste black box & 71.5 & 24.6 & 42.3 & 1333.0 \\
        \quad \quad + paste random noise & 70.5 & 24.2 & 46.7 & 1480.0 \\
    \bottomrule
    \end{tabular}
    }
    \caption{\textbf{Analysis of $i$-CutMix.} Removing weighted loss means that we use one-hot loss instead of the weighted one. We find that simply pasting crops from other images helps greatly, while the weighted loss term can additionally improve the accuracy. We believe the reason is that random crop pasting forces the model to make decisions based on the entire image instead of small patches. We attempt to visualize this in Figure \ref{fig:icutmix_effect}. Note that pasting black box is effectively CutOut augmentation \cite{devries2017improved} and pasting gaussian noise box is effectively RandomErase augmentation \cite{zhong2020random}. Setting: ViT-B, MoCo-v3, poison rate 0.5\%, target category Rottweiler.}
    \label{tab:icutmix_analysis}
\end{table}


\begin{figure}
    \centering
    \includegraphics[width=\columnwidth]{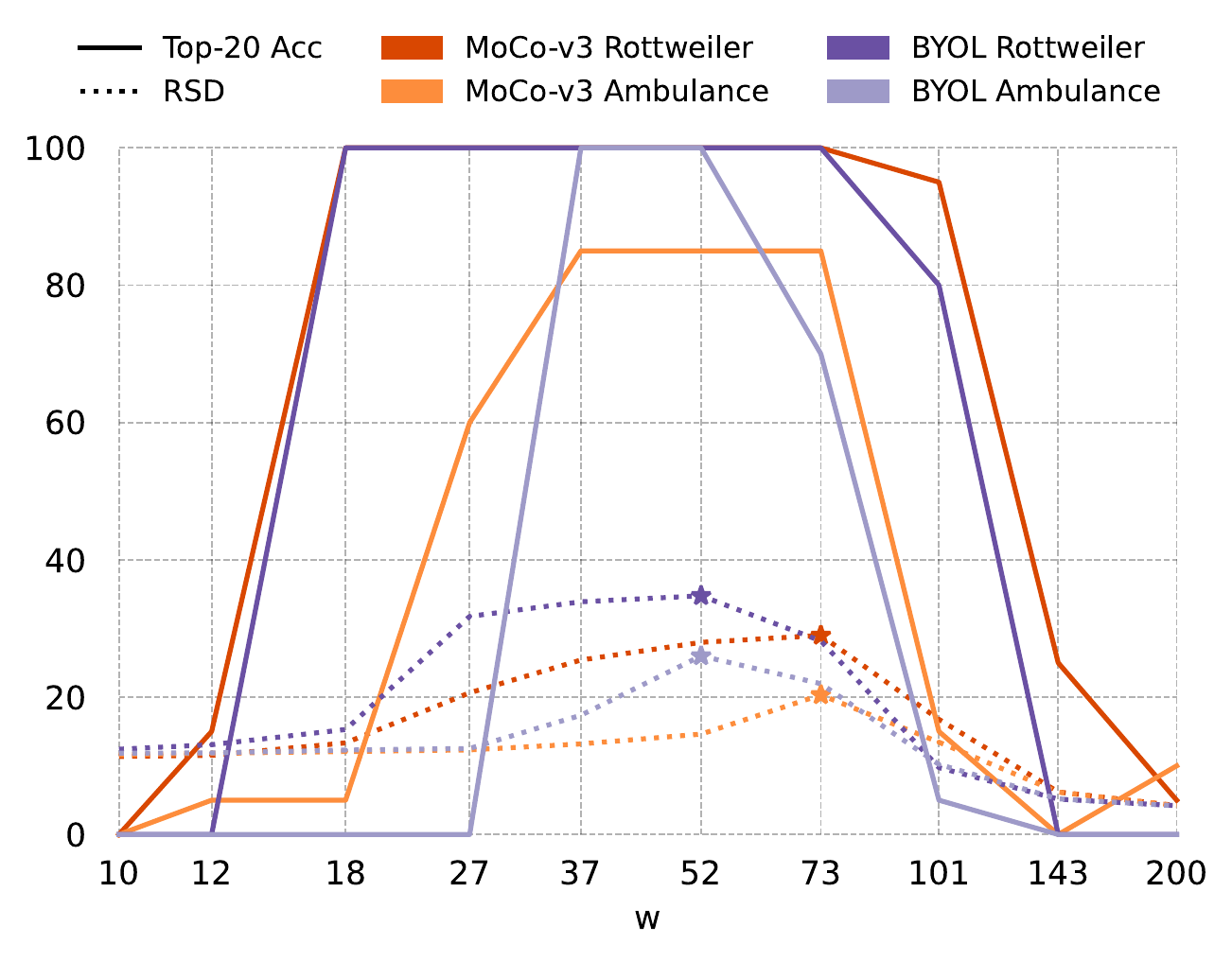}
    \caption{\textbf{Effect of $w$.} We plot the top-$20$ Acc against $w$ values sampled uniformly in log-space from 10 to 200. We find highest accuracies for $w$ near $50$ which is the actual trigger size. However, the defender does not know the actual trigger size. Hence, we propose relative standard deviation (RSD) of the population of poison scores as a heuristic to pick $w$. RSD is standard deviation divided by the mean. In all cases, $w$ with highest RSD (marked with $\filledstar$ in the figure) also corresponds to the highest top-$20$ accuracy.}
    \label{fig:effect_of_w}
\end{figure}

\textbf{Effect of $w$.} An important hyperparameter of PatchSearch is $w$ which controls the size of the extracted candidate trigger. Since the defender does not know the size of the trigger, wrong choice of $w$ can lead to poor poison detection performance as shown in Figure \ref{fig:effect_of_w}. However, it is possible to automate the process of finding a good value for $w$. We know that if the choice of $w$ is wrong, most poison scores will be similar. Hence, the $w$ resulting in high variance in poisons scores should be closer to actual trigger size used by the attacker.
However, we need to adjust the scale of the variance with mean since different $w$ values result in populations with different scales of scores. Hence, we use relative standard deviation (RSD \footnote{\href{https://en.wikipedia.org/wiki/Coefficient_of_variation}{https://en.wikipedia.org/wiki/Coefficient\_of\_variation}}), which is standard deviation divided by the mean. Figure \ref{fig:effect_of_w} shows that picking the $w$ with highest RSD also results in highest top-$20$ accuracy. Therefore, a defender can try different values of $w$ and pick the one with highest RSD without making any assumptions about the trigger size.

\begin{figure}
    \centering
    \includegraphics[width=\columnwidth]{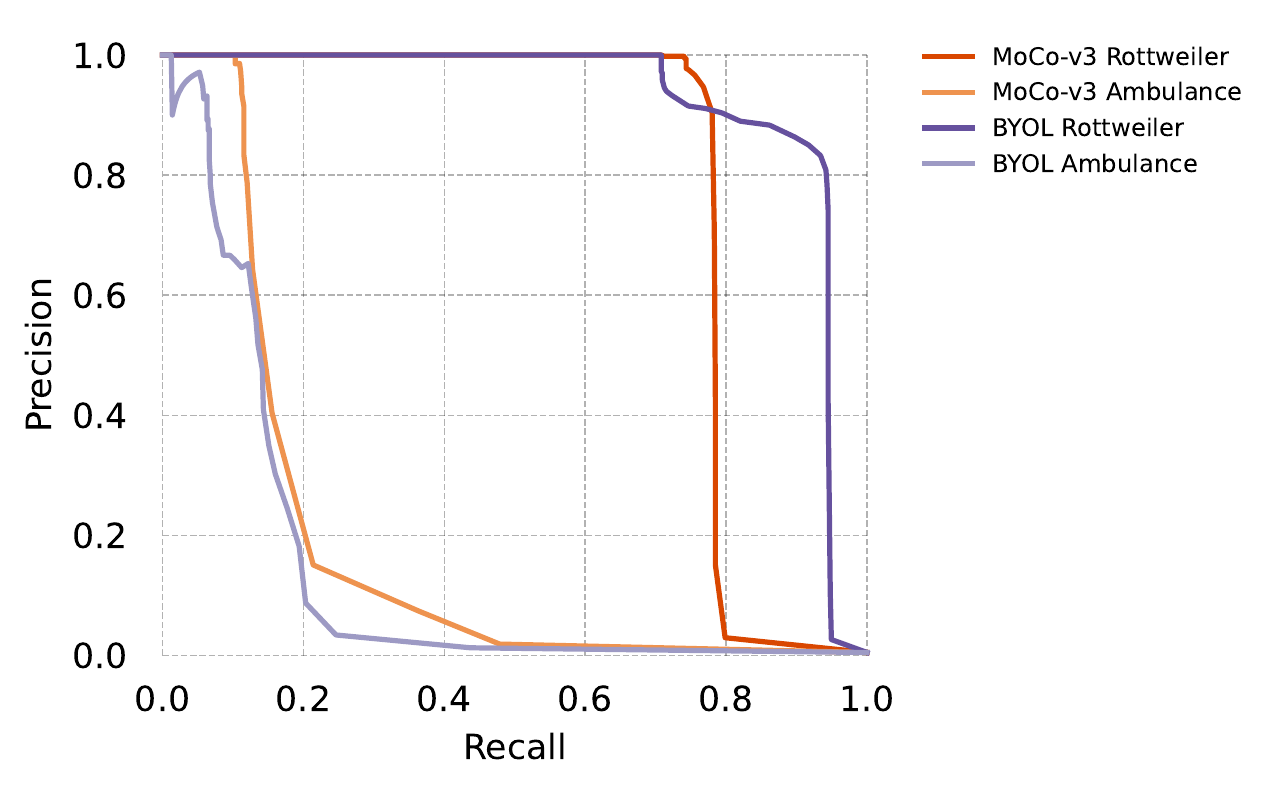}
    \caption{\textbf{Effect of removing poison classifier.} We demonstrate the need for the poison classifier by removing it from PatchSearch, and analyzing the results of images ranked by poison score. Despite scoring every image in the dataset with PatchSearch (no iterative search), poisons cannot be precisely detected in some cases.}
    \label{fig:remove_classifier}
\end{figure}

\textbf{Other ablations.} In Table \ref{tab:repeat_patch_attack}, we explore a stronger form of attack, where the trigger is pasted five times instead of once on target category images in training. This results in the trigger occupying roughly 25\% area of the image. In Table \ref{tab:clean_defense}, we consider the effect of applying PatchSearch to clean data since a defender does not know whether a dataset is poisoned. We see only a small degradation in accuracy. In Figure \ref{fig:remove_classifier}, we consider the effect of removing the poison classifier, and find that resulting image ranking shows poor precision vs. recall trade-off in some cases. See Section \ref{sec:appendix_additional_ablations} of Appendix for more ablations and details. 

\begin{table}
    \centering
    \scalebox{0.90}{
    \begin{tabular}{@{}l cccr@{}}
    \toprule
         & \multicolumn{2}{c}{Clean Data} & \multicolumn{2}{c}{Patched Data} \\
        \cmidrule(lr){2-3} \cmidrule(l){4-5}
        Model & Acc & FP & Acc & FP \\
        \midrule
        Clean & 70.5 & 18.5 & 64.6 & 27.2 \\
        Backdoored & 70.7 & 21.1 & 42.0 & 1996.7 \\
        PatchSearch & 70.3 & 25.2 & 64.7 & 42.0 \\
    \bottomrule
    \end{tabular}
    }
    \caption{\textbf{Repeated patches during attack}. We consider a stronger attack, where five triggers instead of one are pasted on target category images in training. We find that PatchSearch successfully suppresses the attack. Setting: ViT-B, MoCo-v3, poison rate 0.5\%.}
    \label{tab:repeat_patch_attack}
\end{table}

\begin{table}
    \centering
    \scalebox{0.90}{
    \begin{tabular}{@{}l cccr@{}}
    \toprule
          &  \multicolumn{2}{c}{Clean Data} & \multicolumn{2}{c}{Patched Data} \\
        \cmidrule(lr){2-3} \cmidrule(l){4-5}
        Model & Acc & FP & Acc & FP \\
        \midrule
        Clean without PatchSearch & 70.5 & 18.5 & 64.6 & 27.2 \\
        Clean with PatchSearch    & 69.2 & 20.0 & 63.2 & 33.3 \\
    \bottomrule
    \end{tabular}
    }
    \caption{\textbf{Effect of defending a clean model}. Applying PatchSearch on a clean dataset will remove a small set of clean images (6K or about 5\% of the training set), which results in only a small reduction in accuracy. Setting: ViT-B and MoCo-v3.}
    \label{tab:clean_defense}
\end{table}

\section{Conclusion}

We introduced a novel defense algorithm called PatchSearch for defending self-supervised learning against patch based data poisoning backdoor attacks. PatchSearch identifies poisons and removes them from the training set. We also show that the augmentation strategy of $i$-CutMix is a good baseline defense. We find that PatchSearch is better than $i$-CutMix and the SOTA defense that uses trusted data. Further, PatchSearch and $i$-CutMix are complementary to each and their combination improves the model performance while effectively mitigating the attack. However, our defense assumes that the trigger is patch-based and it smaller than the objects of interest. These assumptions may limit its effectiveness against future attacks. We hope that our paper will ecourage the development of better backdoor attacks and defense methods for SSL models.

\noindent \textbf{Acknowledgement.} This work is partially supported by DARPA Contract No. HR00112190135, HR00112290115, and FA8750‐19‐C‐0098, NSF grants 1845216 and 1920079, NIST award 60NANB18D279, and also funding from Shell Inc., and Oracle Corp. 
We would also like to thank K L Navaneet and Aniruddha Saha for many helpful discussions.

{
\small
\bibliographystyle{ieee_fullname}
\bibliography{egbib}
}

\clearpage
\appendix
\section{Appendix}

\setcounter{figure}{0}
\renewcommand\thefigure{A\arabic{figure}}
\setcounter{table}{0}
\renewcommand\thetable{A\arabic{table}}

\begin{figure*}
    \includegraphics[width=\textwidth]{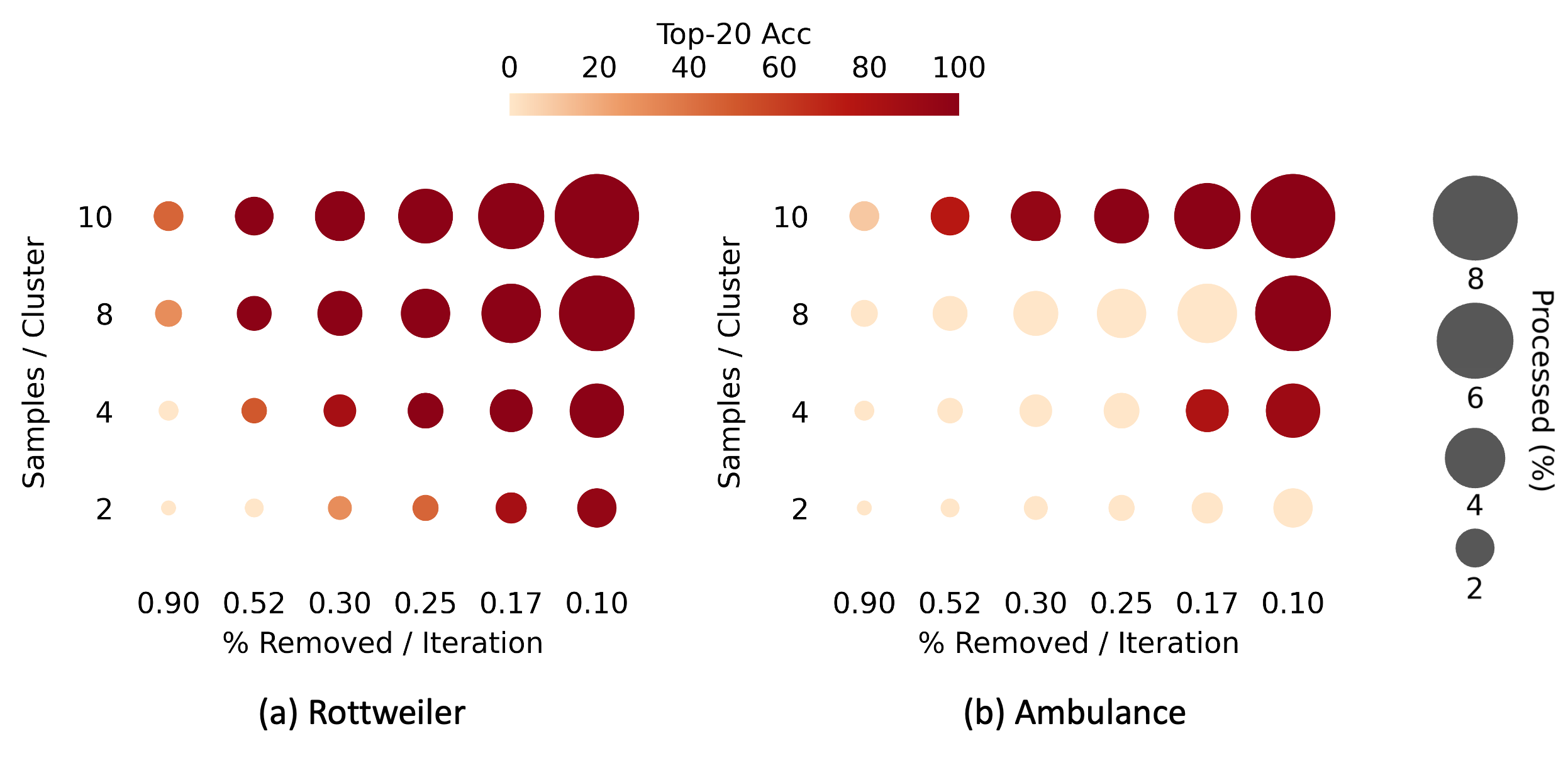}
    \caption{\textbf{Effect of number of images scored.} We vary different hyperparameters that control the number of images scored by PatchSearch and observe the effect on the accuracy of finding poisons in top-$20$ ranked images. The color of the circles denotes the accuracy while their size denotes the number of scored images. We find that a large value for samples per cluster ($s$) has better performance compared to small $s$ values with comparable number of processed images. Finally, processing more than 6\% of images results in a good model performance. See Table \ref{tab:effect_of_processed_count} for detailed results. Seeting: MoCo-v3, ViT-B, and poison rate 0.5\%.}
    \label{fig:effect_of_processed}
\end{figure*}

\begin{table*}[!htbp]
    \centering
    \begin{tabular}{@{} cc rrrrrr @{}}
    \toprule
     Num. & Samples Per &  \multicolumn{6}{c}{Pruned per iteration (\%)} \\
     Clusters & Cluster & \multicolumn{1}{c}{0.90} & \multicolumn{1}{c}{0.52} & \multicolumn{1}{c}{0.30} & \multicolumn{1}{c}{0.25} & \multicolumn{1}{c}{0.17} & \multicolumn{1}{c}{0.10} \\
    \midrule
    100 & 2 & 0.0 / 0.18 & 0.0 / 0.31 & 0.0 / 0.54 & 0.0 / 0.70 & 0.0 / 0.97 & 0.0 / 1.60 \\
    100 & 4 & 0.0 / 0.35 & 0.0 / 0.62 & 0.0 / 1.07 & 0.0 / 1.30 & 80.0 / 1.94 & 90.0 / 3.21 \\
    100 & 8 & 0.0 / 0.70 & 0.0 / 1.24 & 0.0 / 2.14 & 0.0 / 2.60 & 0.0 / 3.87 & 100.0 / 6.42 \\
    100 & 10 & 10.0 / 0.88 & 75.0 / 1.55 & 95.0 / 2.68 & 100.0 / 3.30 & 100.0 / 4.84 & 100.0 / 8.02 \\
    \midrule
    1000 & 2  & 20.0 / 1.80 & 45.0 / 3.00 & 75.0 / 5.30 & 80.0 / 6.40 & 100.0 / 9.40 & 100.0 / 16.00 \\
    1000 & 4  & 65.0 / 3.50 & 100.0 / 6.10 & 100.0 / 10.60 & 95.0 / 12.70 & 100.0 / 18.60 & 100.0 / 30.90 \\
    1000 & 8  & 100.0 / 7.00 & 100.0 / 12.20 & 100.0 / 21.00 & 100.0 / 25.00 & 100.0 / 35.40 & 100.0 / 52.30 \\
    1000 & 10 & 100.0 / 8.80 & 100.0 / 15.20 & 100.0 / 26.10 & 100.0 / 30.80 & 100.0 / 42.40 & 100.0 / 59.30 \\
    \bottomrule
    \end{tabular}
    \caption{\textbf{Effect of varying scored count.} We explore the effect of processing different amount of images by varying number of clusters, $s$ (samples per cluster) and $r$ (percentage of clusters pruned per iteration). Each table entry has the format \textit{accuracy of finding poisons (\%) in top-$20$ / percentage of training set scored}. We find that even with large $r$ and a small number of clusters, increasing $s$ can improve the model performance. Finally, we observe that processing more than 6\% of the training set results in good model performance most of the times. Setting: MoCo-v3, ViT-B, poison rate 0.5\%, target category Ambulance.}
    \label{tab:effect_of_processed_count}
\end{table*}

\begin{table*}
    \centering
    \scalebox{1.0}{
    \begin{tabular}{@{} ccccccc @{}}
    \toprule
    & \multirow{2}{*}{\shortstack{Target\\Category}} & \multicolumn{5}{c}{Flip test set size} \\
    \cmidrule(lr){3-7}
    & & 10 & 32 & 100 & 316 & 1000 \\
    \midrule
    \multirow{2}{*}{top-$20$ Acc (\%)}
    & Ambulance & 5.0 & 20.0 & 60.0 & 80.0 & 80.0 \\
    & Rottweiler & 55.0 & 100.0 & 100.0 & 100.0 & 100.0 \\
    \bottomrule
    \end{tabular}
    }
    \caption{\textbf{Effect of flip test set size}. We explore the effect of changing the flip test set size $|X^f|$. The larger this set the more diverse samples will be used to obtain the poison score for an image. We find that PatchSearch is not greatly sensitive to this hyperparameter and even a small value like 316 could be effective. Setting: ViT-B, MoCo-v3, and poison rate 0.5\%.}
    \label{tab:effect_flip_test_set}
\end{table*}

\begin{table*}
    \centering
    \scalebox{1.0}{
    \begin{tabular}{@{}c rrrr rrrr@{}}
    \toprule
         \multirow{2}{*}{\shortstack{Checkpoint\\Epoch}} & \multicolumn{2}{c}{Clean Data} & \multicolumn{2}{c@{}}{Patched Data} & \multicolumn{2}{c}{Clean Data} & \multicolumn{2}{c@{}}{Patched Data} \\
        \cmidrule(lr){2-3} \cmidrule(l){4-5} \cmidrule(lr){6-7} \cmidrule(l){8-9}
        & Acc & FP & Acc & FP & Acc & FP & Acc & FP \\
        \midrule
 & \multicolumn{4}{c}{\textit{Rottweiler}} & \multicolumn{4}{c}{\textit{Ambulance}} \\
 \cmidrule(lr){2-5} \cmidrule(lr){6-9}
20 & 27.2 & 58.8 & 24.9 & 72.8 & 26.9 & 47.8 & 24.7 & 67.6 \\
40 & 43.6 & 47.4 & 36.4 & 391.8 & 43.3 & 20.6 & 39.6 & 66.2 \\
60 & 53.1 & 36.6 & 17.1 & 3145.0 & 53.0 & 14.0 & 46.5 & 284.6 \\
80 & 58.3 & 32.8 & 35.5 & 1787.4 & 58.0 & 13.4 & 51.2 & 262.4 \\
100 & 60.9 & 28.8 & 28.0 & 2689.6 & 61.4 & 12.6 & 48.0 & 1041.6 \\
120 & 62.9 & 27.6 & 35.5 & 2118.4 & 63.1 & 12.4 & 51.6 & 804.2 \\
140 & 64.2 & 30.0 & 40.7 & 1675.0 & 64.6 & 9.2 & 51.4 & 908.6 \\
160 & 65.8 & 26.4 & 43.5 & 1420.4 & 66.1 & 9.8 & 54.8 & 704.6 \\
180 & 66.3 & 25.8 & 40.9 & 1895.8 & 66.9 & 7.8 & 54.4 & 800.6 \\
200 & 66.4 & 26.0 & 39.9 & 1926.8 & 66.8 & 7.6 & 53.6 & 895.6 \\
    \bottomrule
    \end{tabular}
    }
    \caption{\textbf{Relationship between overall performance and attack effectiveness}. We evaluate intermediate checkpoints to understand the relationship between overall model performance (measured with \textit{Clean Data Acc}) and the attack effectiveness (measured with \textit{Patched Data FP}). We find that attack effectiveness has a strong correlation with overall model performance in early epochs ($<= 60$). While this correlation is not strict for later epochs, the attack effectiveness maintains its overall magnitude. This observation suggests that a model used during defense need not be trained to convergence. We only need to train it until the attack effectiveness is comparable to the fully trained model in overall magnitude. Setting: BYOL, ResNet-18, and poison rate 0.5\%.}
    \label{tab:checkpoint_eval}
\end{table*}

\begin{table}
    \centering
    \scalebox{1.0}{
    \begin{tabular}{@{}l ccccrr@{}}
    \toprule
          &  \multicolumn{3}{c}{Clean Data} & \multicolumn{3}{c}{Patched Data} \\
        \cmidrule(lr){2-4} \cmidrule(l){5-7}
        Model & Acc & FP & ASR & Acc & FP & ASR \\
        \midrule
        Clean & 79.2 & 205 & 2.1 & 54.3 & 741 & 7.4 \\
        Backdoored & 79.0 & 206 & 2.1 & 26.5 & 6187 & 61.9 \\
        PatchSearch & 78.7 & 224 & 2.2 & 57.0 & 816 & 8.2 \\
    \bottomrule
    \end{tabular}
    }
    \caption{\textbf{CIFAR-10}. We show results for the attack and defense on CIFAR-10. We find that the defense is successfully able to mitigate the attack. In fact, we find that the poison classifier is not needed since the iterative search itself is sufficient. Simply removing the top 10\% of the clusters removes 100\% of poisons. To account for smaller, 32x32, images, we also reduce the size of the trigger to 8x8. Setting: ResNet-18, BYOL, poison rate 0.5\% and target category Airplane.}
    \label{tab:cifar10}
\end{table}

\begin{table}
    \centering
    \scalebox{0.85}{
    \begin{tabular}{@{}l ccccc@{}}
    \toprule
        & \multicolumn{5}{c}{Acc (\%) in top-$k$}\\
        Model & 5 & 10 & 20 & 50 & 100 \\ 
        \midrule
        ResNet-18, BYOL, 0.5\% & 100.0 & 100.0 & 99.5 & 93.8 & 84.0 \\ 
        ResNet-18, MoCo-v2, 0.5\% & 52.0 & 55.0 & 52.5 & 44.6 & 30.9 \\ 
        ViT-B, MoCo-v3, 0.5\%  & 100.0 & 100.0 & 96.5 & 87.6 & 73.5 \\ 
        ViT-B, MoCo-v3, 1.0\%  & 98.0 & 98.0 & 97.5 & 84.0 & 77.3 \\ 
    \bottomrule
    \end{tabular}
    }
    \caption{\textbf{Effect of top-$k$ in PatchSearch.} Note that PatchSearch is not able to find the patches for ResNet-18, MoCo-v2 for any $k$. Hence, we use an easy to backdoor model like ViT-B to defend it.}
    \label{tab:effect_of_k}
\end{table}

\begin{table}
    \centering
    \scalebox{1.0}{
    \begin{tabular}{@{}l ccccrr@{}}
    \toprule
          &  \multicolumn{3}{c}{Clean Data} & \multicolumn{3}{c}{Patched Data} \\
        \cmidrule(lr){2-4} \cmidrule(l){5-7}
        Model & Acc & FP & ASR & Acc & FP & ASR \\
        \midrule
        \multicolumn{4}{l}{$w=75$} \\
        Clean & 65.7 & 28.8 & 0.6 & 55.0 & 19.0 & 0.4 \\
        Backdoored & 66.6 & 32.4 & 0.7 & 36.4 & 1588.0 & 32.0 \\
        PatchSearch & 66.5 & 29.2 & 0.6 & 55.8 & 25.2 & 0.5 \\
        \midrule
        \multicolumn{4}{l}{$w=100$} \\
        Clean & 65.7 & 28.8 & 0.6 & 45.6 & 15.2 & 0.3 \\
        Backdoored & 65.8 & 30.2 & 0.6 & 20.4 & 2481.6 & 50.6 \\
        PatchSearch & 65.6 & 32.8 & 0.7 & 46.7 & 34.6 & 0.7 \\
    \bottomrule
    \end{tabular}
    }
    \caption{\textbf{Changing trigger size $w$ in the attack.}. We explore the effect of changing the trigger size $w$ during the attack. We find that our defense can successfully defend against different $w$ by searching for most effective $w$ during the defense (outlined in Figure \ref{fig:effect_of_w}). Note that since we are dealing with large $w$ we paste the trigger anywhere in the image without any boundary. This is different from the default setting of 25\% margin on all sides. Setting: ResNet-18, BYOL, poison rate 0.5\% and target category Rottweiler.}
\label{tab:effect_trigger_size}
\end{table}


\begin{table}
    \centering
    \scalebox{0.85}{
    \begin{tabular}{@{}ll rrrr@{}}
    \toprule
        \multirow{2}{*}{\shortstack{Target\\Category}} & \multirow{2}{*}{\shortstack{Model}} & \multicolumn{2}{c}{Clean Data} & \multicolumn{2}{c@{}}{Patched Data} \\
        \cmidrule(lr){3-4} \cmidrule(l){5-6}
        & & Acc & FP & Acc & FP \\
        \midrule
        \multirow{3}{*}{Chicken Curry (10)}
        & Clean & 47.6 & 27 & 39.1 & 24 \\
        & Backdoored & 48.2 & 19 & 9.2 & 3438 \\
        & PatchSearch & 46.6 & 29 & 40.3 & 21 \\
        \midrule
        \multirow{3}{*}{Steak (11)}
        & Clean & 47.6 & 17 & 39.5 & 24 \\
        & Backdoored & 48.0 & 26 & 13.5 & 2596 \\
        & PatchSearch & 47.4 & 22 & 40.7 & 57 \\
        \midrule
        \multirow{3}{*}{Panna Cotta (12)}
        & Clean & 47.6 & 41 & 39.8 & 23 \\
        & Backdoored & 47.5 & 42 & 19.1 & 2158 \\
        & PatchSearch & 47.4 & 37 & 40.9 & 16 \\
        \midrule
        \multirow{3}{*}{Deviled Eggs (13)}
        & Clean & 47.6 & 50 & 39.3 & 41 \\
        & Backdoored & 47.9 & 23 & 7.9 & 4317 \\
        & PatchSearch & 48.0 & 52 & 41.0 & 35 \\
        \midrule
        \multirow{3}{*}{\textit{Mean}}
        & Clean & 47.6 & 33.8 & 39.4 & 28.0 \\
        & Backdoored & 47.9 & 27.5 & 12.4 & 3127.3 \\
        & PatchSearch & 47.4 & 35.0 & 40.7 & 32.3 \\
    \bottomrule
    \end{tabular}
    }
    \caption{\textbf{Attack on downstream task: Food101 classification}. Instead of attacking one of the categories in the pre-training dataset, we consider an attack on the downstream task of Food101 classification \cite{food101}. We poison 700 images of a category from Food101 and add them to the pre-training dataset (ImageNet-100). We evaluate the resulting models on Food101 classification and find that the targeted attack is successful. Next, we defend the backdoored models with PatchSearch and find that it is able to defend against the attack successfully. Averaged across the four categories, PatchSearch has 99.7\% recall and 54.8\% precision for filtering out poisons. Setting: BYOL \& ResNet-18.}
    \label{tab:proper_downstream_appendix}
\end{table}

\subsection{Additional Ablations}
\label{sec:appendix_additional_ablations}

We consider the effect of number of images scored by iterative search in Table \ref{tab:effect_of_processed_count} and Figure \ref{fig:effect_of_processed}. Number of scored images can be changed by varying two hyperparameters $s$ (number of samples scored per cluster) and $r$ (percentage of least poisonous clusters removed in each iteration). The results show consistently good performance when processing more than 6\% of samples. Next, we consider the effect of varying the size of flip test set $X^f$ in Table \ref{tab:effect_flip_test_set}. We find that we can get reasonable results with $|X^f|$ as small as 316. Next, we evaluate intermediate checkpoints of one of our models to understand the relationship between overall performance of the model (measured with clean data Acc) and the attack effectiveness (measured with patched data FP) in Table \ref{tab:checkpoint_eval}. We can see that attack effectiveness improves as the overall model performance improves in the early epochs. In later epochs, the attack effectiveness fluctuates but is still high relative to earlier epochs. This indicates that it is possible to reduce the compute requirements for the model used during defense. We only need to train this model until the attack effectiveness becomes broadly comparable to a fully trained model. Next, we show that our defense is effective even when changing datasets in Table \ref{tab:cifar10}. Next, we consider the effect of varying top-$k$ in Table \ref{tab:effect_of_k}, and find that any top-$k <= 20$ is a good choice. Next, we consider changing the trigger size $w$ during the attack in Table \ref{tab:effect_trigger_size}. We find that our defense works despite changing $w$ from 50 to 75 and 100. Next, we consider attacking images from a downstream dataset instead of pre-training dataset. We use Food101 \cite{food101} classification dataset for this purpose. It consists of 101 fine-grained food categories and 750 images per category. For each category, we randomly select 700 images for training and 50 images for validation. We backdoor BYOL, ResNet-18 models by picking random category from Food101, poisoning all of its 700 training images, and adding them to the pre-training dataset (ImageNet-100). The resulting pre-trained model is evaluated on the task of Food101 classification. We use the same linear evaluation setup as other ResNet-18 models in our experiments. We use the training set of Food101 ($700 \times 101$) for training the linear layer while the evaluation set ($50 \times 101$) is used for evaluation. The results are presented in Table \ref{tab:proper_downstream_appendix} We find that the attack is successful, but the models can also be successfully defended with PatchSearch.



\subsection{Implementation Details}
\label{sec:appendix_implementation_details}


Below, we describe implementation details for various settings. Note that running PatchSearch is relatively inexpensive compared to the pre-training stage. For instance, with ImageNet-100 (126K images) and ViT-B on 4x3090 GPUs, PatchSearch takes 1.5 hrs, which is small as compared to the training time, about 16 hrs, for MoCo-v3.

\textbf{Poison classifier training in PatchSearch.} The training has following parameters: SGD (lr=$0.01$, batch size=$32$, max iterations=$2000$, weight decay=$1e^{-4}$, and cosine lr scheduler). The architecture of the classifier is ResNet-18 but each layer has only a single \texttt{BasicBlock} instead of the default two \footnote{\url{https://github.com/pytorch/vision/blob/main/torchvision/models/resnet.py}}.

\textbf{ResNet-18 model training.} As mentioned in the main paper, MoCo-v2 and BYOL training for ResNet-18 models is exactly the same as code  \footnote{\url{https://github.com/UMBCvision/SSL-Backdoor}} from [41]. The models are trained on 4 NVIDIA A100 GPUs.

\textbf{MoCo-v3 model training.} We use the code \footnote{\url{https://github.com/facebookresearch/moco-v3}} from MoCo-v3 [11] for training ViT-B models. We use the default hyperparameters from the code except SGD (batch size=$1024$, epochs=$200$). The models are trained on 8 NVIDIA A100 GPUs.

\textbf{MAE model training.} We use the code \footnote{\url{https://github.com/facebookresearch/mae}} from MAE [23] to train ViT-B models. All hyperparameters are unchanged except following: SGD (batch size=$32$ and
accum iter=$4$). Here, \textit{accum iter} refers to the number iterations used for averaging the gradients before updating parameters. The models are trained on 8 NVIDIA A100 GPUs. Hence, the effective batch size is $32\times8\times4=1024$.

\textbf{ResNet-18 linear evaluation.} We use the linear layer training procedure proposed in code \footnote{\url{https://github.com/UMBCvision/CompRess/blob/master/eval\_linear.py}} from ComPress [4] for evaluating ResNet-18 models. A single linear layer is trained on top of a frozen backbone. The output of the backbone is processed according to following steps before passing it to the linear layer. (1) A mini-batch of features is normalized to have unit $l_2$ norm. (2) The mini-batch is also normalized to have zero mean and unit variance. Note that the mean and variance used for normalization in the second step comes from the $l_2$ normalized features of the entire training dataset. All hyperparameters values are set to default values from the original code.

\textbf{ViT-B linear evaluation.} We use the linear layer training procedure from MoCo-v3 [11] code \footnote{\url{https://github.com/facebookresearch/moco-v3/blob/main/main\_lincls.py}}. We set all hyperparameters to their default values from the code except SGD (epochs=$30$, batch size=$256$).

\textbf{ViT-B fine-tuning evaluation.} We use the code \footnote{\url{https://github.com/facebookresearch/mae/blob/main/engine\_finetune.py}} from MAE \cite{he2022masked} for fine-tuning ViT-B models. Strong augmentations like mixup, random erase, and cutmix are turned off during fine-tuning since we find that their presence hurts the overall performance of the model (Table \ref{tab:finetune_strong_aug}). For MoCo-v3 models, fine-tuning runs for $30$ epochs while for MAE models it runs for $90$ epochs. Finally, MAE uses global pooling of tokens following the default option in the original code while MoCo-v3 models use \texttt{[CLS]} token. Rest of the hyperparameters are unchanged from their default values. For fine-tuning results on ImageNetx-1k reported in Table \ref{tab:moco_v3_vs_mae_official_finetune}, all settings except epochs=$50$ are the same as their default values. Note that strong augmentations is kept on as in the default settings in order to be comparable to the numbers reported in MAE [23].

\begin{table}[t]
    \centering
    \scalebox{1.0}{
    \begin{tabular}{@{}l cccr@{}}
    \toprule
          &  \multicolumn{2}{c}{Clean Data} & \multicolumn{2}{c}{Patched Data} \\
        \cmidrule(lr){2-3} \cmidrule(l){4-5}
        Model & Acc & FP & Acc & FP \\
        \midrule
        with strong aug & 63.1 & 21.0 & 53.5 & 63.1 \\
        without strong aug & 65.7 & 18.7 & 53.8 & 97.6 \\
    \bottomrule
    \end{tabular}
    }
    \caption{\textbf{Effect of strong augmentations during fine-tuning}. We find that strong augmentations like cutmix, mixup, and random erase can degrade the overall model performance. Setting: ViT-B, MAE and average of 4 target categories (Rottweiler, Tabby Cat, Ambulance, and Laptop).}
    \label{tab:finetune_strong_aug}
\end{table}

\subsection{Per Category Results}
\label{sec:appendix_per_category_results}

We list per category results for Table \ref{tab:main_defense_results}. The results are in Tables \ref{tab:byol_resnet18_rate_0x5}, \ref{tab:moco_v2_resnet18_rate_0x5}, \ref{tab:moco_v3_vit_b_rate_0x5}, and \ref{tab:moco_v3_vit_b_rate_1x0}.

\begin{table*}
    \centering
    \scalebox{0.70}{
    \begin{tabular}{@{}ll cc rrrr@{}}
    \toprule
        Target & & & & \multicolumn{2}{c}{Clean Data} & \multicolumn{2}{c}{Patched Data} \\
        \cmidrule(lr){5-6} \cmidrule(l){7-8}
        Category & Model & PatchSearch & $i$-CutMix & Acc & FP & Acc & FP \\
        \midrule
\multirow{6}{*}{Rottweiler (10)} 
 & clean & \xmark & \xmark & 65.7 & 29.4 & 60.6 & 24.4 \\
 & clean & \xmark & \cmark & 65.8 & 33.0 & 64.0 & 28.6 \\
 & backdoored & \xmark & \xmark & 66.4 & 26.0 & 39.9 & 1926.8 \\
 & defended & \xmark & \cmark & 66.8 & 26.4 & 60.4 & 272.0 \\
 & defended & \cmark & \xmark & 66.9 & 31.8 & 61.1 & 43.8 \\
 & defended & \cmark & \cmark & 65.6 & 32.6 & 63.4 & 32.2 \\
\midrule
\multirow{6}{*}{Tabby Cat (11)} 
 & clean & \xmark & \xmark & 65.7 & 5.0 & 60.8 & 9.4 \\
 & clean & \xmark & \cmark & 65.8 & 4.4 & 63.8 & 3.4 \\
 & backdoored & \xmark & \xmark & 65.9 & 1.6 & 31.9 & 2338.0 \\
 & defended & \xmark & \cmark & 67.1 & 6.4 & 62.5 & 299.0 \\
 & defended & \cmark & \xmark & 66.1 & 4.2 & 61.9 & 7.0 \\
 & defended & \cmark & \cmark & 67.0 & 3.2 & 64.6 & 1.6 \\
\midrule
\multirow{6}{*}{Ambulance (12)} 
 & clean & \xmark & \xmark & 65.7 & 8.6 & 60.5 & 10.2 \\
 & clean & \xmark & \cmark & 65.8 & 9.2 & 63.9 & 10.4 \\
 & backdoored & \xmark & \xmark & 66.8 & 7.6 & 53.6 & 895.6 \\
 & defended & \xmark & \cmark & 66.5 & 7.8 & 62.4 & 116.0 \\
 & defended & \cmark & \xmark & 66.4 & 9.2 & 61.2 & 9.2 \\
 & defended & \cmark & \cmark & 67.3 & 7.2 & 65.1 & 7.0 \\
\midrule
\multirow{6}{*}{Pickup Truck (13)} 
 & clean & \xmark & \xmark & 65.7 & 14.2 & 60.6 & 13.2 \\
 & clean & \xmark & \cmark & 65.8 & 14.2 & 63.7 & 15.6 \\
 & backdoored & \xmark & \xmark & 66.4 & 13.6 & 48.0 & 1430.6 \\
 & defended & \xmark & \cmark & 68.3 & 16.8 & 64.1 & 131.0 \\
 & defended & \cmark & \xmark & 65.7 & 14.8 & 60.8 & 13.0 \\
 & defended & \cmark & \cmark & 67.1 & 19.0 & 64.6 & 18.8 \\
\midrule
\multirow{6}{*}{Laptop (14)} 
 & clean & \xmark & \xmark & 65.7 & 30.8 & 60.4 & 33.0 \\
 & clean & \xmark & \cmark & 65.8 & 27.4 & 64.1 & 15.8 \\
 & backdoored & \xmark & \xmark & 65.6 & 29.6 & 16.1 & 355.8 \\
 & defended & \xmark & \cmark & 66.2 & 25.2 & 60.5 & 356.2 \\
 & defended & \cmark & \xmark & 65.8 & 30.4 & 60.9 & 37.6 \\
 & defended & \cmark & \cmark & 66.3 & 29.0 & 63.9 & 17.0 \\
\midrule
\multirow{6}{*}{Goose (15)} 
 & clean & \xmark & \xmark & 65.7 & 10.4 & 60.3 & 18.0 \\
 & clean & \xmark & \cmark & 65.8 & 8.6 & 63.6 & 9.0 \\
 & backdoored & \xmark & \xmark & 66.4 & 10.0 & 26.5 & 3391.0 \\
 & defended & \xmark & \cmark & 67.4 & 11.2 & 62.4 & 288.2 \\
 & defended & \cmark & \xmark & 66.4 & 12.2 & 61.6 & 23.6 \\
 & defended & \cmark & \cmark & 67.0 & 25.6 & 65.1 & 24.8 \\
\midrule
\multirow{6}{*}{Pirate Ship (16)} 
 & clean & \xmark & \xmark & 65.7 & 4.2 & 60.4 & 4.6 \\
 & clean & \xmark & \cmark & 65.8 & 5.4 & 63.5 & 6.0 \\
 & backdoored & \xmark & \xmark & 66.9 & 4.6 & 38.3 & 2438.6 \\
 & defended & \xmark & \cmark & 67.2 & 4.4 & 61.4 & 328.2 \\
 & defended & \cmark & \xmark & 66.4 & 4.0 & 61.5 & 9.8 \\
 & defended & \cmark & \cmark & 66.9 & 3.6 & 64.4 & 2.4 \\
\midrule
\multirow{6}{*}{Gas Mask (17)} 
 & clean & \xmark & \xmark & 65.7 & 19.8 & 60.7 & 28.6 \\
 & clean & \xmark & \cmark & 65.8 & 22.0 & 63.8 & 36.4 \\
 & backdoored & \xmark & \xmark & 66.6 & 21.2 & 34.8 & 2722.4 \\
 & defended & \xmark & \cmark & 66.8 & 18.6 & 57.2 & 931.6 \\
 & defended & \cmark & \xmark & 66.8 & 30.8 & 62.1 & 52.8 \\
 & defended & \cmark & \cmark & 67.2 & 29.0 & 64.9 & 55.8 \\
\midrule
\multirow{6}{*}{Vacuum Cleaner (18)} 
 & clean & \xmark & \xmark & 65.7 & 60.0 & 60.6 & 74.8 \\
 & clean & \xmark & \cmark & 65.8 & 58.6 & 63.4 & 38.2 \\
 & backdoored & \xmark & \xmark & 66.6 & 49.2 & 20.4 & 3210.0 \\
 & defended & \xmark & \cmark & 67.0 & 51.0 & 59.6 & 328.4 \\
 & defended & \cmark & \xmark & 66.8 & 58.4 & 61.8 & 85.8 \\
 & defended & \cmark & \cmark & 67.0 & 59.0 & 64.8 & 39.2 \\
\midrule
\multirow{6}{*}{American Lobseter (19)} 
 & clean & \xmark & \xmark & 65.7 & 24.0 & 61.0 & 31.4 \\
 & clean & \xmark & \cmark & 65.8 & 21.8 & 63.5 & 32.6 \\
 & backdoored & \xmark & \xmark & 67.0 & 22.2 & 43.5 & 2086.8 \\
 & defended & \xmark & \cmark & 66.5 & 17.6 & 59.5 & 602.2 \\
 & defended & \cmark & \xmark & 66.3 & 32.8 & 61.9 & 51.0 \\
 & defended & \cmark & \cmark & 66.6 & 15.8 & 64.2 & 26.6 \\
\midrule
			
\multirow{6}{*}{Mean and STD} 
 & clean & \xmark & \xmark & 65.7 \small{$\pm$  0.0} & 20.6 \small{$\pm$ 16.8} & 60.6 \small{$\pm$  0.2} & 24.8 \small{$\pm$ 20.2} \\
 & clean & \xmark & \cmark & 65.8 \small{$\pm$  0.0} & 20.5 \small{$\pm$ 16.5} & 63.7 \small{$\pm$  0.2} & 19.6 \small{$\pm$ 13.1} \\
 & backdoored & \xmark & \xmark & 66.5 \small{$\pm$  0.4} & 18.6 \small{$\pm$ 14.3} & 35.3 \small{$\pm$ 11.9} & 2079.6 \small{$\pm$ 967.5} \\
 & defended & \xmark & \cmark & 67.0 \small{$\pm$  0.6} & 18.5 \small{$\pm$ 13.7} & 61.0 \small{$\pm$  2.0} & 365.3 \small{$\pm$ 239.4} \\
 & defended & \cmark & \xmark & 66.4 \small{$\pm$  0.4} & 22.9 \small{$\pm$ 17.1} & 61.5 \small{$\pm$  0.5} & 33.4 \small{$\pm$ 25.6} \\
 & defended & \cmark & \cmark & 66.8 \small{$\pm$  0.5} & 22.4 \small{$\pm$ 16.8} & 64.5 \small{$\pm$  0.5} & 22.5 \small{$\pm$ 17.1} \\
    \bottomrule
    \end{tabular}
    }
    \caption{\textbf{Per category results for BYOL, ResNet-18, and poison rate 0.5\%}. Detailed results for Table \ref{tab:main_defense_results}.}
    \label{tab:byol_resnet18_rate_0x5}
\end{table*}

\begin{table*}
    \centering
    \scalebox{0.70}{
    \begin{tabular}{@{}ll cc cccr@{}}
    \toprule
        Target & & & & \multicolumn{2}{c}{Clean Data} & \multicolumn{2}{c}{Patched Data} \\
        \cmidrule(lr){5-6} \cmidrule(l){7-8}
        Category & Model & PatchSearch & $i$-CutMix & Acc & FP & Acc & FP \\
        \midrule
\multirow{6}{*}{Rottweiler (10)} 
 & clean & \xmark & \xmark & 49.7 & 36.8 & 46.3 & 28.4 \\
 & clean & \xmark & \cmark & 55.9 & 32.0 & 54.0 & 29.4 \\
 & backdoored & \xmark & \xmark & 50.2 & 39.4 & 33.4 & 1094.4 \\
 & defended & \xmark & \cmark & 55.1 & 38.6 & 52.4 & 117.2 \\
 & defended & \cmark & \xmark & 50.0 & 40.2 & 46.3 & 26.6 \\
 & defended & \cmark & \cmark & 55.6 & 39.0 & 54.4 & 35.2 \\
\midrule
\multirow{6}{*}{Tabby Cat (11)} 
 & clean & \xmark & \xmark & 49.7 & 7.2 & 46.4 & 8.2 \\
 & clean & \xmark & \cmark & 55.9 & 8.0 & 54.4 & 6.6 \\
 & backdoored & \xmark & \xmark & 50.0 & 5.8 & 33.5 & 1901.6 \\
 & defended & \xmark & \cmark & 55.3 & 5.6 & 52.4 & 181.8 \\
 & defended & \cmark & \xmark & 49.9 & 11.6 & 46.7 & 13.8 \\
 & defended & \cmark & \cmark & 55.7 & 8.8 & 54.2 & 9.0 \\
\midrule
\multirow{6}{*}{Ambulance (12)} 
 & clean & \xmark & \xmark & 49.7 & 18.4 & 46.4 & 15.0 \\
 & clean & \xmark & \cmark & 55.9 & 13.6 & 54.4 & 19.2 \\
 & backdoored & \xmark & \xmark & 50.3 & 14.0 & 46.2 & 103.2 \\
 & defended & \xmark & \cmark & 55.4 & 10.6 & 53.8 & 27.6 \\
 & defended & \cmark & \xmark & 49.0 & 14.4 & 45.4 & 15.0 \\
 & defended & \cmark & \cmark & 55.7 & 16.4 & 54.1 & 15.8 \\
\midrule
\multirow{6}{*}{Pickup Truck (13)} 
 & clean & \xmark & \xmark & 49.7 & 16.6 & 46.6 & 15.4 \\
 & clean & \xmark & \cmark & 55.9 & 16.6 & 54.2 & 18.6 \\
 & backdoored & \xmark & \xmark & 50.6 & 13.0 & 46.4 & 115.0 \\
 & defended & \xmark & \cmark & 55.8 & 15.2 & 53.8 & 75.2 \\
 & defended & \cmark & \xmark & 49.1 & 18.2 & 45.8 & 17.8 \\
 & defended & \cmark & \cmark & 55.5 & 16.6 & 53.9 & 18.6 \\
\midrule
\multirow{6}{*}{Laptop (14)} 
 & clean & \xmark & \xmark & 49.7 & 37.0 & 46.0 & 35.8 \\
 & clean & \xmark & \cmark & 55.9 & 33.0 & 54.3 & 24.2 \\
 & backdoored & \xmark & \xmark & 49.8 & 33.8 & 41.8 & 466.2 \\
 & defended & \xmark & \cmark & 55.4 & 24.0 & 53.7 & 92.6 \\
 & defended & \cmark & \xmark & 49.7 & 41.6 & 45.8 & 47.6 \\
 & defended & \cmark & \cmark & 54.8 & 46.4 & 53.5 & 29.4 \\
\midrule
\multirow{6}{*}{Goose (15)} 
 & clean & \xmark & \xmark & 49.7 & 37.2 & 46.6 & 39.4 \\
 & clean & \xmark & \cmark & 55.9 & 38.2 & 54.2 & 39.2 \\
 & backdoored & \xmark & \xmark & 49.6 & 33.8 & 45.2 & 194.0 \\
 & defended & \xmark & \cmark & 55.5 & 31.0 & 53.8 & 46.8 \\
 & defended & \cmark & \xmark & 49.5 & 39.6 & 46.3 & 46.0 \\
 & defended & \cmark & \cmark & 55.1 & 41.0 & 53.4 & 37.6 \\
\midrule
\multirow{6}{*}{Pirate Ship (16)} 
 & clean & \xmark & \xmark & 49.7 & 8.2 & 46.1 & 9.0 \\
 & clean & \xmark & \cmark & 55.9 & 5.0 & 54.4 & 6.4 \\
 & backdoored & \xmark & \xmark & 49.7 & 6.2 & 42.1 & 573.4 \\
 & defended & \xmark & \cmark & 55.7 & 4.6 & 53.6 & 68.8 \\
 & defended & \cmark & \xmark & 49.7 & 12.8 & 46.8 & 17.4 \\
 & defended & \cmark & \cmark & 55.0 & 5.6 & 53.3 & 8.2 \\
\midrule
\multirow{6}{*}{Gas Mask (17)} 
 & clean & \xmark & \xmark & 49.7 & 43.8 & 46.5 & 57.0 \\
 & clean & \xmark & \cmark & 55.9 & 39.6 & 54.2 & 56.8 \\
 & backdoored & \xmark & \xmark & 49.7 & 43.6 & 42.3 & 561.2 \\
 & defended & \xmark & \cmark & 55.2 & 37.2 & 53.2 & 92.6 \\
 & defended & \cmark & \xmark & 50.0 & 47.4 & 46.6 & 62.4 \\
 & defended & \cmark & \cmark & 55.6 & 48.6 & 53.8 & 61.8 \\
\midrule
\multirow{6}{*}{Vacuum Cleaner (18)} 
 & clean & \xmark & \xmark & 49.7 & 51.4 & 46.5 & 63.2 \\
 & clean & \xmark & \cmark & 55.9 & 53.6 & 54.3 & 40.0 \\
 & backdoored & \xmark & \xmark & 49.8 & 48.8 & 41.7 & 424.4 \\
 & defended & \xmark & \cmark & 55.4 & 54.2 & 53.7 & 55.6 \\
 & defended & \cmark & \xmark & 49.9 & 58.0 & 46.4 & 64.6 \\
 & defended & \cmark & \cmark & 54.8 & 51.2 & 53.4 & 37.0 \\
\midrule
\multirow{6}{*}{American Lobseter (19)} 
 & clean & \xmark & \xmark & 49.7 & 34.0 & 46.5 & 35.4 \\
 & clean & \xmark & \cmark & 55.9 & 23.4 & 54.1 & 30.4 \\
 & backdoored & \xmark & \xmark & 49.8 & 34.0 & 36.1 & 1599.2 \\
 & defended & \xmark & \cmark & 55.2 & 23.2 & 49.7 & 482.2 \\
 & defended & \cmark & \xmark & 50.0 & 47.2 & 46.7 & 61.2 \\
 & defended & \cmark & \cmark & 55.6 & 32.2 & 54.3 & 39.2 \\
\midrule
			
\multirow{6}{*}{Mean and STD} 
 & clean & \xmark & \xmark & 49.7 \small{$\pm$  0.0} & 29.1 \small{$\pm$ 15.3} & 46.4 \small{$\pm$  0.2} & 30.7 \small{$\pm$ 19.2} \\
 & clean & \xmark & \cmark & 55.9 \small{$\pm$  0.0} & 26.3 \small{$\pm$ 15.6} & 54.3 \small{$\pm$  0.1} & 27.1 \small{$\pm$ 15.6} \\
 & backdoored & \xmark & \xmark & 50.0 \small{$\pm$  0.3} & 27.2 \small{$\pm$ 16.0} & 40.9 \small{$\pm$  4.9} & 703.3 \small{$\pm$ 626.1} \\
 & defended & \xmark & \cmark & 55.4 \small{$\pm$  0.2} & 24.4 \small{$\pm$ 16.1} & 53.0 \small{$\pm$  1.3} & 124.0 \small{$\pm$ 132.9} \\
 & defended & \cmark & \xmark & 49.7 \small{$\pm$  0.4} & 33.1 \small{$\pm$ 17.1} & 46.3 \small{$\pm$  0.5} & 37.2 \small{$\pm$ 21.3} \\
 & defended & \cmark & \cmark & 55.3 \small{$\pm$  0.4} & 30.6 \small{$\pm$ 17.2} & 53.8 \small{$\pm$  0.4} & 29.2 \small{$\pm$ 16.6} \\

    \bottomrule
    \end{tabular}
    }
    \caption{\textbf{Per category results for MoCo-v2, ResNet-18, and poison rate 0.5\%}.Detailed results for Table \ref{tab:main_defense_results}.}
    \label{tab:moco_v2_resnet18_rate_0x5}
\end{table*}

\begin{table*}
    \centering
    \scalebox{0.70}{
    \begin{tabular}{@{}ll cc cccr@{}}
    \toprule
        Target & & & & \multicolumn{2}{c}{Clean Data} & \multicolumn{2}{c}{Patched Data} \\
        \cmidrule(lr){5-6} \cmidrule(l){7-8}
        Category & Model & PatchSearch & $i$-CutMix & Acc & FP & Acc & FP \\
        \midrule
\multirow{6}{*}{Rottweiler (10)} 
 & clean & \xmark & \xmark & 70.5 & 25.4 & 65.1 & 16.6 \\
 & clean & \xmark & \cmark & 75.6 & 31.0 & 74.5 & 26.8 \\
 & backdoored & \xmark & \xmark & 70.5 & 24.8 & 42.9 & 1412.4 \\
 & defended & \xmark & \cmark & 75.6 & 28.0 & 70.5 & 268.4 \\
 & defended & \cmark & \xmark & 70.4 & 28.2 & 64.8 & 23.8 \\
 & defended & \cmark & \cmark & 75.7 & 32.6 & 74.9 & 28.6 \\
\midrule
\multirow{6}{*}{Tabby Cat (11)} 
 & clean & \xmark & \xmark & 70.5 & 3.2 & 64.4 & 4.0 \\
 & clean & \xmark & \cmark & 75.6 & 5.0 & 74.5 & 3.4 \\
 & backdoored & \xmark & \xmark & 70.7 & 4.6 & 42.0 & 2354.6 \\
 & defended & \xmark & \cmark & 75.4 & 5.2 & 73.3 & 133.4 \\
 & defended & \cmark & \xmark & 70.6 & 6.0 & 65.2 & 7.6 \\
 & defended & \cmark & \cmark & 74.7 & 4.8 & 73.8 & 4.4 \\
\midrule
\multirow{6}{*}{Ambulance (12)} 
 & clean & \xmark & \xmark & 70.5 & 9.8 & 64.4 & 13.4 \\
 & clean & \xmark & \cmark & 75.6 & 8.2 & 74.4 & 8.0 \\
 & backdoored & \xmark & \xmark & 70.5 & 10.2 & 56.9 & 748.8 \\
 & defended & \xmark & \cmark & 75.3 & 8.6 & 74.4 & 49.6 \\
 & defended & \cmark & \xmark & 70.1 & 9.6 & 63.8 & 16.6 \\
 & defended & \cmark & \cmark & 75.2 & 6.8 & 74.1 & 8.6 \\
\midrule
\multirow{6}{*}{Pickup Truck (13)} 
 & clean & \xmark & \xmark & 70.5 & 13.8 & 65.3 & 10.4 \\
 & clean & \xmark & \cmark & 75.6 & 11.0 & 74.3 & 13.0 \\
 & backdoored & \xmark & \xmark & 70.7 & 11.8 & 57.9 & 805.0 \\
 & defended & \xmark & \cmark & 75.7 & 12.4 & 74.3 & 54.2 \\
 & defended & \cmark & \xmark & 69.9 & 14.0 & 65.9 & 12.8 \\
 & defended & \cmark & \cmark & 75.1 & 13.0 & 74.3 & 13.6 \\
\midrule
\multirow{6}{*}{Laptop (14)} 
 & clean & \xmark & \xmark & 70.5 & 29.6 & 64.9 & 39.4 \\
 & clean & \xmark & \cmark & 75.6 & 34.4 & 74.5 & 26.6 \\
 & backdoored & \xmark & \xmark & 70.5 & 34.6 & 42.3 & 2172.4 \\
 & defended & \xmark & \cmark & 75.8 & 31.4 & 71.7 & 365.8 \\
 & defended & \cmark & \xmark & 69.8 & 43.0 & 62.9 & 93.2 \\
 & defended & \cmark & \cmark & 75.3 & 40.8 & 74.3 & 31.0 \\
\midrule
\multirow{6}{*}{Goose (15)} 
 & clean & \xmark & \xmark & 70.5 & 6.4 & 64.8 & 9.6 \\
 & clean & \xmark & \cmark & 75.6 & 6.6 & 74.4 & 6.8 \\
 & backdoored & \xmark & \xmark & 70.7 & 6.8 & 48.6 & 1693.2 \\
 & defended & \xmark & \cmark & 76.1 & 5.2 & 74.0 & 66.6 \\
 & defended & \cmark & \xmark & 69.6 & 10.0 & 64.8 & 17.4 \\
 & defended & \cmark & \cmark & 75.1 & 7.2 & 74.0 & 5.0 \\
\midrule
\multirow{6}{*}{Pirate Ship (16)} 
 & clean & \xmark & \xmark & 70.5 & 2.0 & 64.2 & 1.4 \\
 & clean & \xmark & \cmark & 75.6 & 2.6 & 74.1 & 1.6 \\
 & backdoored & \xmark & \xmark & 70.7 & 2.2 & 56.2 & 915.6 \\
 & defended & \xmark & \cmark & 75.8 & 2.0 & 73.5 & 93.4 \\
 & defended & \cmark & \xmark & 70.4 & 2.0 & 62.1 & 1.6 \\
 & defended & \cmark & \cmark & 75.7 & 1.6 & 74.6 & 1.2 \\
\midrule
\multirow{6}{*}{Gas Mask (17)} 
 & clean & \xmark & \xmark & 70.5 & 29.0 & 64.5 & 55.6 \\
 & clean & \xmark & \cmark & 75.6 & 12.4 & 74.3 & 23.0 \\
 & backdoored & \xmark & \xmark & 70.3 & 20.0 & 39.2 & 2558.0 \\
 & defended & \xmark & \cmark & 75.4 & 14.2 & 71.0 & 381.6 \\
 & defended & \cmark & \xmark & 70.4 & 42.6 & 64.6 & 71.2 \\
 & defended & \cmark & \cmark & 74.9 & 28.0 & 73.8 & 44.2 \\
\midrule
\multirow{6}{*}{Vacuum Cleaner (18)} 
 & clean & \xmark & \xmark & 70.5 & 52.0 & 64.4 & 113.8 \\
 & clean & \xmark & \cmark & 75.6 & 36.2 & 74.5 & 25.4 \\
 & backdoored & \xmark & \xmark & 70.6 & 49.8 & 43.8 & 1847.4 \\
 & defended & \xmark & \cmark & 75.2 & 35.4 & 66.3 & 723.4 \\
 & defended & \cmark & \xmark & 70.4 & 60.8 & 65.2 & 128.0 \\
 & defended & \cmark & \cmark & 75.0 & 49.4 & 73.8 & 36.2 \\
\midrule
\multirow{6}{*}{American Lobseter (19)} 
 & clean & \xmark & \xmark & 70.5 & 13.6 & 64.5 & 8.0 \\
 & clean & \xmark & \cmark & 75.6 & 8.2 & 74.3 & 11.4 \\
 & backdoored & \xmark & \xmark & 70.7 & 9.4 & 39.0 & 2581.8 \\
 & defended & \xmark & \cmark & 75.9 & 6.4 & 73.3 & 185.4 \\
 & defended & \cmark & \xmark & 70.8 & 14.4 & 65.7 & 25.8 \\
 & defended & \cmark & \cmark & 75.1 & 12.4 & 74.2 & 16.8 \\
\midrule
			
\multirow{6}{*}{Mean and STD} 
 & clean & \xmark & \xmark & 70.5 \small{$\pm$  0.0} & 18.5 \small{$\pm$ 15.6} & 64.6 \small{$\pm$  0.4} & 27.2 \small{$\pm$ 34.8} \\
 & clean & \xmark & \cmark & 75.6 \small{$\pm$  0.0} & 15.6 \small{$\pm$ 13.0} & 74.4 \small{$\pm$  0.1} & 14.6 \small{$\pm$  10.0} \\
 & backdoored & \xmark & \xmark & 70.6 \small{$\pm$  0.1} & 17.4 \small{$\pm$ 15.1} & 46.9 \small{$\pm$  7.5} & 1708.9 \small{$\pm$ 714.2} \\
 & defended & \xmark & \cmark & 75.6 \small{$\pm$  0.3} & 14.9 \small{$\pm$ 12.2} & 72.2 \small{$\pm$  2.5} & 232.2 \small{$\pm$ 212.5} \\
 & defended & \cmark & \xmark & 70.2 \small{$\pm$  0.4} & 23.1 \small{$\pm$ 19.6} & 64.5 \small{$\pm$  1.2} & 39.8 \small{$\pm$ 42.6} \\
 & defended & \cmark & \cmark & 75.2 \small{$\pm$  0.3} & 19.7 \small{$\pm$ 16.8} & 74.2 \small{$\pm$  0.4} & 19.0 \small{$\pm$ 15.0} \\
    \bottomrule
    \end{tabular}
    }
    \caption{\textbf{Per category results for MoCo-v3, ViT-B, and poison rate 0.5\%}. Detailed results for Table \ref{tab:main_defense_results}.}
    \label{tab:moco_v3_vit_b_rate_0x5}
\end{table*}

\begin{table*}
    \centering
    \scalebox{0.70}{
    \begin{tabular}{@{}ll cc cccr@{}}
    \toprule
        Target & & & & \multicolumn{2}{c}{Clean Data} & \multicolumn{2}{c}{Patched Data} \\
        \cmidrule(lr){5-6} \cmidrule(l){7-8}
        Category & Model & PatchSearch & $i$-CutMix & Acc & FP & Acc & FP \\
        \midrule
\multirow{6}{*}{Rottweiler (10)} 
 & clean & \xmark & \xmark & 70.5 & 25.4 & 65.1 & 16.6 \\
 & clean & \xmark & \cmark & 75.6 & 31.0 & 74.5 & 26.8 \\
 & backdoored & \xmark & \xmark & 70.8 & 21.8 & 30.6 & 3127.4 \\
 & defended & \xmark & \cmark & 75.6 & 21.4 & 69.0 & 510.8 \\
 & defended & \cmark & \xmark & 70.0 & 32.6 & 64.8 & 26.8 \\
 & defended & \cmark & \cmark & 75.3 & 37.4 & 74.3 & 32.0 \\
\midrule
\multirow{6}{*}{Tabby Cat (11)} 
 & clean & \xmark & \xmark & 70.5 & 3.2 & 64.4 & 4.0 \\
 & clean & \xmark & \cmark & 75.6 & 5.0 & 74.5 & 3.4 \\
 & backdoored & \xmark & \xmark & 70.9 & 3.4 & 23.8 & 3669.8 \\
 & defended & \xmark & \cmark & 75.7 & 5.0 & 71.0 & 429.6 \\
 & defended & \cmark & \xmark & 70.4 & 24.4 & 62.9 & 86.8 \\
 & defended & \cmark & \cmark & 74.8 & 29.6 & 74.0 & 18.0 \\
\midrule
\multirow{6}{*}{Ambulance (12)} 
 & clean & \xmark & \xmark & 70.5 & 9.8 & 64.4 & 13.4 \\
 & clean & \xmark & \cmark & 75.6 & 8.2 & 74.4 & 8.0 \\
 & backdoored & \xmark & \xmark & 70.7 & 7.4 & 49.0 & 1511.2 \\
 & defended & \xmark & \cmark & 75.4 & 7.4 & 73.0 & 180.8 \\
 & defended & \cmark & \xmark & 69.8 & 25.8 & 64.2 & 36.0 \\
 & defended & \cmark & \cmark & 75.0 & 21.0 & 73.9 & 21.0 \\
\midrule
\multirow{6}{*}{Pickup Truck (13)} 
 & clean & \xmark & \xmark & 70.5 & 13.8 & 65.3 & 10.4 \\
 & clean & \xmark & \cmark & 75.6 & 11.0 & 74.3 & 13.0 \\
 & backdoored & \xmark & \xmark & 70.7 & 12.4 & 54.6 & 1007.0 \\
 & defended & \xmark & \cmark & 75.2 & 9.6 & 73.0 & 117.8 \\
 & defended & \cmark & \xmark & 69.7 & 18.4 & 64.9 & 19.2 \\
 & defended & \cmark & \cmark & 75.2 & 18.6 & 73.9 & 21.4 \\
\midrule
\multirow{6}{*}{Laptop (14)} 
 & clean & \xmark & \xmark & 70.5 & 29.6 & 64.9 & 39.4 \\
 & clean & \xmark & \cmark & 75.6 & 34.4 & 74.5 & 26.6 \\
 & backdoored & \xmark & \xmark & 70.8 & 27.0 & 39.4 & 2566.4 \\
 & defended & \xmark & \cmark & 75.4 & 25.8 & 66.9 & 724.8 \\
 & defended & \cmark & \xmark & 69.9 & 61.2 & 60.3 & 112.8 \\
 & defended & \cmark & \cmark & 75.6 & 54.2 & 74.7 & 55.0 \\
\midrule
\multirow{6}{*}{Goose (15)} 
 & clean & \xmark & \xmark & 70.5 & 6.4 & 64.8 & 9.6 \\
 & clean & \xmark & \cmark & 75.6 & 6.6 & 74.4 & 6.8 \\
 & backdoored & \xmark & \xmark & 70.5 & 5.8 & 38.6 & 2437.8 \\
 & defended & \xmark & \cmark & 75.7 & 5.2 & 71.2 & 314.2 \\
 & defended & \cmark & \xmark & 70.1 & 32.8 & 63.4 & 59.0 \\
 & defended & \cmark & \cmark & 74.7 & 33.4 & 73.7 & 25.0 \\
\midrule
\multirow{6}{*}{Pirate Ship (16)} 
 & clean & \xmark & \xmark & 70.5 & 2.0 & 64.2 & 1.4 \\
 & clean & \xmark & \cmark & 75.6 & 2.6 & 74.1 & 1.6 \\
 & backdoored & \xmark & \xmark & 70.3 & 2.6 & 44.9 & 2093.8 \\
 & defended & \xmark & \cmark & 75.8 & 2.4 & 73.5 & 90.8 \\
 & defended & \cmark & \xmark & 70.3 & 20.2 & 62.4 & 36.0 \\
 & defended & \cmark & \cmark & 75.0 & 18.2 & 74.0 & 24.2 \\
\midrule
\multirow{6}{*}{Gas Mask (17)} 
 & clean & \xmark & \xmark & 70.5 & 29.0 & 64.5 & 55.6 \\
 & clean & \xmark & \cmark & 75.6 & 12.4 & 74.3 & 23.0 \\
 & backdoored & \xmark & \xmark & 70.4 & 21.2 & 31.6 & 3112.0 \\
 & defended & \xmark & \cmark & 76.1 & 8.8 & 67.7 & 728.6 \\
 & defended & \cmark & \xmark & 70.2 & 79.8 & 64.7 & 138.8 \\
 & defended & \cmark & \cmark & 74.6 & 66.2 & 73.5 & 102.8 \\
\midrule
\multirow{6}{*}{Vacuum Cleaner (18)} 
 & clean & \xmark & \xmark & 70.5 & 52.0 & 64.4 & 113.8 \\
 & clean & \xmark & \cmark & 75.6 & 36.2 & 74.5 & 25.4 \\
 & backdoored & \xmark & \xmark & 71.2 & 38.4 & 40.6 & 2203.4 \\
 & defended & \xmark & \cmark & 75.3 & 30.6 & 64.9 & 903.0 \\
 & defended & \cmark & \xmark & 70.4 & 85.8 & 64.8 & 182.8 \\
 & defended & \cmark & \cmark & 74.6 & 69.2 & 73.8 & 63.2 \\
\midrule
\multirow{6}{*}{American Lobseter (19)} 
 & clean & \xmark & \xmark & 70.5 & 13.6 & 64.5 & 8.0 \\
 & clean & \xmark & \cmark & 75.6 & 8.2 & 74.3 & 11.4 \\
 & backdoored & \xmark & \xmark & 70.3 & 8.0 & 24.1 & 3630.2 \\
 & defended & \xmark & \cmark & 75.1 & 9.0 & 71.6 & 341.6 \\
 & defended & \cmark & \xmark & 70.0 & 53.4 & 64.3 & 61.6 \\
 & defended & \cmark & \cmark & 75.1 & 42.2 & 74.3 & 51.0 \\
\midrule
			
\multirow{6}{*}{Mean and STD}
 & clean & \xmark & \xmark & 70.5 \small{$\pm$  0.0} & 18.5 \small{$\pm$ 15.6} & 64.6 \small{$\pm$  0.4} & 27.2 \small{$\pm$ 34.8} \\
 & clean & \xmark & \cmark & 75.6 \small{$\pm$  0.0} & 15.6 \small{$\pm$ 13.0} & 74.4 \small{$\pm$  0.1} & 14.6 \small{$\pm$  10.0} \\
 & backdoored & \xmark & \xmark & 70.7 \small{$\pm$  0.3} & 14.8 \small{$\pm$ 11.8} & 37.7 \small{$\pm$  10.2} & 2535.9 \small{$\pm$ 873.6} \\
 & defended & \xmark & \cmark & 75.5 \small{$\pm$  0.3} & 12.5 \small{$\pm$ 9.8} & 70.2 \small{$\pm$  2.9} & 434.2 \small{$\pm$ 279.3} \\
 & defended & \cmark & \xmark & 70.1 \small{$\pm$  0.2} & 43.4 \small{$\pm$ 25.0} & 63.7 \small{$\pm$  1.5} & 76.0 \small{$\pm$ 53.9} \\
 & defended & \cmark & \cmark & 75.0 \small{$\pm$  0.3} & 39.0 \small{$\pm$ 18.8} & 74.0 \small{$\pm$  0.3} & 41.4 \small{$\pm$ 27.0} \\
    \bottomrule
    \end{tabular}
    }
    \caption{\textbf{Per category results for MoCo-v3, ViT-B, and poison rate 1.0\%}. Detailed results for Table \ref{tab:main_defense_results}.}
    \label{tab:moco_v3_vit_b_rate_1x0}
\end{table*}

\end{document}